\pdfoutput=1

\documentclass[11pt]{article}

\usepackage[preprint]{acl}

\usepackage{times}
\usepackage{latexsym}

\usepackage[T1]{fontenc}

\usepackage[utf8]{inputenc}

\usepackage{microtype}

\usepackage{inconsolata}

\usepackage{xspace}
\usepackage{algorithm}
\usepackage{algpseudocode}
\usepackage{tikz}
\usepackage{fancybox}
\usepackage{booktabs}
\usepackage{multirow} 
\usepackage{threeparttable}
\usepackage{tabularx}
\usepackage{ragged2e}
\usepackage{enumitem}
\setlist{nolistsep}

\newcommand{\MLLMs}[0]{MLLMs\xspace}
\newcommand{\MLLMsLong}[0]{Multimodal Large Language Models\xspace}
\newcommand{\MLLM}[0]{MLLM\xspace}

\newcommand{\ICL}[0]{ICL\xspace}
\newcommand{\ICLLong}[0]{In-Context-Learning\xspace}

\newcommand{\CoT}[0]{CoT\xspace}
\newcommand{\CoTLong}[0]{Chain-of-Thought\xspace}

\title{Evaluating Linguistic Capabilities of Multimodal LLMs\\in the Lens of Few-Shot Learning}

\author{Mustafa Dogan\textsuperscript{1,2}\thanks{Corresponding author, \href{mailto:dogankaas@gmail.com}{{dogan\_mustafa@hacettepe.edu.tr}}}
\quad Ilker Kesen\textsuperscript{3,4}
\quad \textbf{Iacer Calixto\textsuperscript{5,6}}
\quad \textbf{Aykut Erdem\textsuperscript{3,4}}
\quad \textbf{Erkut Erdem\textsuperscript{1,3}}
\vspace{0.2cm}\\
\textsuperscript{1} Hacettepe University, Department of Computer Engineering~ \textsuperscript{2} Aselsan Research \\
\textsuperscript{3} Ko\c{c} University, KUIS AI Center~ \textsuperscript{4} Ko\c{c} University, Department of Computer Engineering\\
\textsuperscript{5} Amsterdam UMC, University of Amsterdam, Department of Medical Informatics \\
\textsuperscript{6} Amsterdam Public Health, Methodology \& Mental Health, Amsterdam, The Netherlands\\
}

\begin{document}
\maketitle
\begin{abstract}

The linguistic capabilities of Multimodal Large Language Models (MLLMs) are critical for their effective application across diverse tasks. This study aims to evaluate the performance of MLLMs on the VALSE benchmark, focusing on the efficacy of few-shot In-Context Learning (ICL), and Chain-of-Thought (CoT) prompting. We conducted a comprehensive assessment of state-of-the-art MLLMs, varying in model size and pretraining datasets. The experimental results reveal that ICL and CoT prompting significantly boost model performance, particularly in tasks requiring complex reasoning and contextual understanding. Models pretrained on captioning datasets show superior zero-shot performance, while those trained on interleaved image-text data benefit from few-shot learning. Our findings provide valuable insights into optimizing MLLMs for better grounding of language in visual contexts, highlighting the importance of the composition of pretraining data and the potential of few-shot learning strategies to improve the reasoning abilities of MLLMs.

\end{abstract}

\section{Introduction} \label{sec:introduction}

\MLLMsLong (\MLLMs) demonstrate a remarkable ability to interpret both text and other modalities, such as images \citep{chen2022pali, alayrac2022flamingo, tsimpoukelli2021multimodal, awadalla2023openflamingo, laurenccon2023obelics, Li2023OtterAM}. These models integrate visual and textual data, allowing them to perform a wide range of reasoning tasks effectively.
Despite their impressive capabilities, optimizing these models through fine-tuning is resource-intensive and costly. To address these challenges, researchers have developed efficient data augmentation techniques and optimization algorithms \citep{huang2018multimodal, falcon2020data, mou2020multimodal}. Among these, few-shot learning techniques offer a promising solution by significantly reducing the costs associated with fine-tuning \citep{chen2023understanding, tsimpoukelli2021multimodal, wei2022chain, wang2022self}.

\begin{figure*}[!t]
\includegraphics[width=\linewidth]{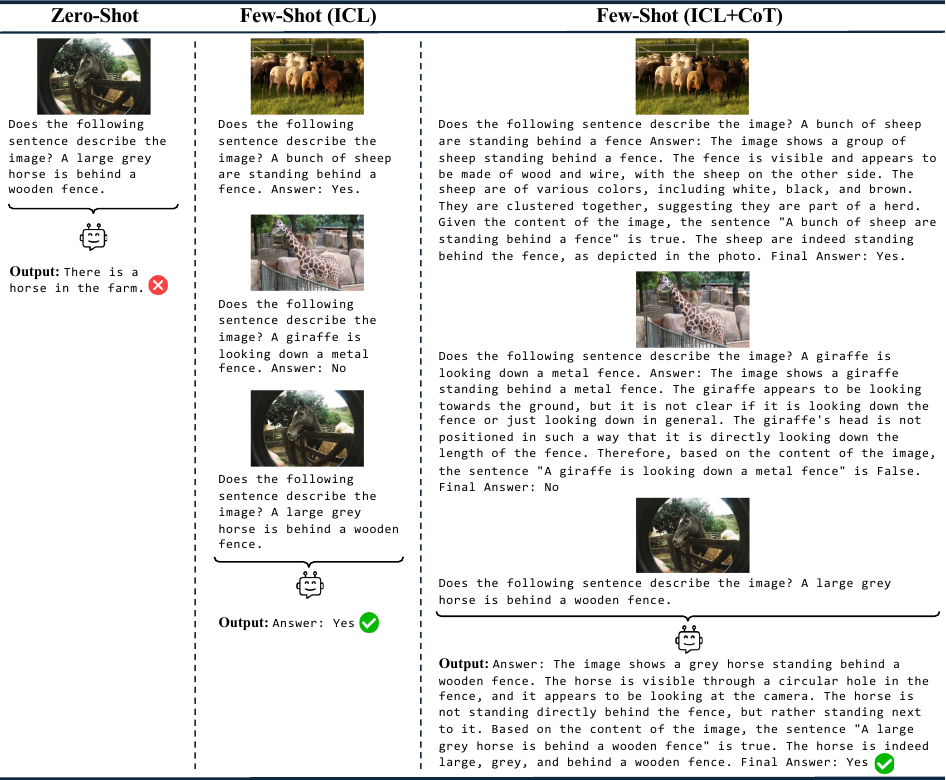}

    \caption{\textbf{Zero-Shot, Few-Shot ICL, and Few-Shot ICL+CoT Evaluation Comparison on the \emph{Relations} Task.} 
In the Zero-Shot approach, the model incorrectly responds to the question. Few-Shot ICL, using prior examples, correctly identifies the horse behind a wooden fence. Few-Shot ICL+CoT, which is beneficial for tasks requiring intermediate reasoning steps, e.g. counting, relational understanding, and coreference resolution, also correctly identifies the horse by employing a detailed step-by-step reasoning process.}
    \label{fig:icl-definition}
\end{figure*}

Few-shot learning is an \ICLLong (\ICL) strategy that enhances model performance by providing a small number of demonstration examples, introducing a specific context \citep{brown2020language}. This method allows the model to leverage its inherent knowledge, combined with the context provided, to solve complex tasks in various domains without specific prior training. %
\CoTLong (\CoT) \citep{wei2022chain} is, on the other hand, a prompting methodology which involves generating reasoning chains before providing the final answer. This strategy enables models to produce more accurate outputs, especially for tasks that require intermediate steps and reasoning, such as arithmetic and commonsense reasoning. 
Without these reasoning chains, models often fail when they respond with only the final answer.

Although the individual effects of few-shot \ICL and \CoT strategies have been studied in multimodal settings, their collective impact on the linguistic capabilities of \MLLMs has not been previously explored. Understanding this impact is crucial, as it can significantly influence the development and deployment of more efficient and capable \MLLMs. To address this gap, we utilize the VALSE (Vision And Language Structured Evaluation) benchmark \citep{parcalabescu-etal-2022-valse}. VALSE provides a comprehensive framework for evaluating the grounding of linguistic phenomena—from morphosyntax to semantics—in the visual modality. It includes six tasks: Existence, Plurality, Counting, Spatial Relations, Actions, and Coreference. These tasks are designed to test models' abilities to recognize existential quantifiers, semantic number, entity counting, spatial arrangements, actions, and pronominal coreference within images. 

The VALSE benchmark is particularly well-suited for this study because it functions as a test-only benchmark without any training data, which aligns perfectly with the ICL setting of our analysis. This allows us to evaluate the models purely based on their pre-existing capabilities and the provided context, without any additional fine-tuning or training. VALSE offers a detailed understanding of how models handle specific linguistic constructs and tasks, highlighting their strengths and deficiencies in visio-linguistic grounding. This makes it an ideal choice for examining the collective impact of ICL and CoT on the linguistic capabilities of~MLLMs.

Using VALSE, we aim to investigate the effects of \ICL and \CoT on the performance of \MLLMs. Our study makes the following contributions:
\begin{itemize}[leftmargin=*]
\item %
We conduct a thorough evaluation of 14 different \MLLMs on VALSE. This evaluation examines both zero-shot and few-shot settings, providing insights into how demonstration examples and reasoning chains influence model outputs.
\item %
Our results indicate that using demonstration examples in the few-shot \ICL setting enhances overall performance. Notably, examples similar to the query image-text pairs significantly boost performance compared to randomly selected examples, as in prior work~\citep{liu-etal-2022-makes,luo2023dricl}.
\item %
\CoT proves highly effective for tasks requiring intermediate reasoning steps, such as counting, relational understanding, and coreference resolution. This highlights the potential of CoT in enhancing the reasoning capabilities of MLLMs.
\item %
We demonstrate that models pretrained on captioning datasets such as MS-COCO \citep{lin2014microsoft}, Conceptual Captions \citep{sharma-etal-2018-conceptual}, and LAION-5B \citep{schuhmann2022laion5b} exhibit superior zero-shot performance compared to those trained on interleaved image-text datasets like Multimodal C4 \citep{zhu2023multimodal} and OBELISC \citep{laurenccon2023obelics}. However, with few-shot ICL strategies, lower-capacity models trained on interleaved image-text datasets can achieve similar or even better performance than the larger-capacity models trained on captioning datasets.
\end{itemize}

The subsequent sections of this paper are organized as follows: In \S \ref{sec:relatedWorks}, we provide a concise review of relevant literature. \S \ref{sec:methodology} outlines our evaluation strategy, offering comprehensive insights into our approach. In \S \ref{sec:experiments}, we present our results. \S \ref{sec:conclusion} gives our conclusions, summarizing the key findings and implications derived from this study. Lastly, in  \S\ref{sec:limitations}, we share the limitations of our study.

\section{Related Work} \label{sec:relatedWorks}

In this section, we will explore the specifics of the recent \MLLMs (\S\ref{sec:relatedWorks:pretrainedILMs}), current \ICL and \CoT techniques (\S\ref{sec:relatedWorks:fewShotLearning} and \S{\ref{sec:relatedWorks:CoT}}), examining their evolution, applications, and emerging approaches in this rapidly developing area.

\subsection{\MLLMsLong} \label{sec:relatedWorks:pretrainedILMs}
\noindent \textbf{Pretraining Strategies.} \MLLMsLong (\MLLMs) require different pretraining datasets to support various capabilities. \MLLMs often use datasets of image-text pairs due to several advantages: they are easy to use, provide a direct relationship between text and image, and include well-established, widely-used, and standardized datasets \citep{lin2014microsoft, plummer2016flickr30k, schuhmann2022laion5b, changpinyo2021cc12m}. Conversely, interleaved image-text datasets \citep{zhu2023multimodal, laurenccon2023obelics, li2023mimic, zhao2023mmicl} create a context with multiple images and texts, enabling models to leverage this context to solve complex tasks. This approach allows models to tackle new challenges, such as narrating a series of images. Additionally, instruction-tuning datasets \citep{liu2024visual, chen2023sharegpt4v, li2023mimic} are crucial for enhancing the flexibility and responsiveness of these models. By training on a diverse set of instructions paired with corresponding outputs, these datasets enable models to follow specific prompts more accurately and generalize better across different tasks. This improves the models' capabilities in zero-shot and few-shot learning scenarios, making them more versatile and effective for real-world applications where diverse and precise responses are needed. 

\noindent \textbf{Models.} The development of \MLLMs has significantly advanced, leveraging the capabilities of pre-trained autoregressive LLMs and sophisticated visual encoders to handle both text and visual inputs \citep{chen2023internvl, dong2024internlm, zhu2023minigpt, fuyu-8b}. Notable examples include Flamingo \citep{alayrac2022flamingo}, which has demonstrated remarkable performance across various vision-language tasks. This progress has led to the creation of open-weight models, fostering collaboration and accessibility in the field \citep{ye2023mplug, Li2023OtterAM, Emu2, lu2024deepseek, jiang2024mantis, awadalla2023openflamingo, xgen_mm_phi3_mini, zhao2023mmicl}. IDEFICS models\citep{laurençon2024matters, laurenccon2023obelics} surpasses inference efficiency and stable training by leveraging pre-trained unimodal backbones. Similarly, Qwen-VL Chat \citep{Qwen-VL}, based on Qwen-7B, emphasizes fine-grained visual understanding and multilingual support, achieving state-of-the-art performance. In contrast, LLaVA-NeXT \citep{liu2024llavanext}, an improved version of LLaVA-1.5 \citep{liu2023improvedllava}, employs a surprisingly powerful and data-efficient vision-language integration module, requiring only training a simple fully-connected projection layer on a modest dataset. While Qwen-VL trains specially designed visual resamplers on vast amounts of image-text paired data, LLaVA-NeXT achieves SOTA results with publicly available data, demonstrating efficiency and effectiveness in model design and training. MMICL \citep{zhao2023mmicl} addresses limitations in current models by efficiently handling multi-modal inputs, including relationships among multiple images and text-to-image references. By introducing a novel context scheme and a comprehensive multi-modal \ICL dataset, MMICL significantly improves understanding of intricate text-image relationships and multi-image reasoning.

\subsection{\ICLLong (\ICL)} 
\label{sec:relatedWorks:fewShotLearning}
\ICL was first developed for LLMs, where the goal is to provide a context with examples that the model can use to solve complex tasks \cite{brown2020language}. To transfer \ICL for \MLLMs, researchers train these models using interleaved image-text datasets. Selecting demonstration examples for \ICL is critical, and the multimodal nature of \MLLMs makes this selection more challenging, as it requires finding examples that are appropriate both textually and visually. Some studies suggest choosing examples based on their similarity to the query image-text pair \citep{alayrac2022flamingo, chen2023understanding, gui2021kat, lin2022revive, liu2021makes}. However, research \citep{shukor2023beyond} indicates that \ICL can increase hallucinations and has a limited impact on improving image-text matching and instruction-following abilities. Additionally, \citealt{chen2023understanding} found that while image similarity has a slight effect on model performance in Visual Question Answering (VQA) tasks, it raises questions about the overall effectiveness of \ICL in multimodal settings. Several recent studies have begun to explore the In-Context Learning (ICL) capabilities of MLLMs. \citet{shukor2023beyond} examined the impact of ICL, Chain-of-Hindsight ICL \citep{liu2023}, and Self-Correcting ICL \citep{madaan2023} on factors such as hallucinations, abstention, compositionality, explainability, and instruction following. \citet{zhao2023mmicl} evaluated the effect of ICL on the performance of a few MLLMs using standard vision-language datasets. In contrast, our study provides a more comprehensive analysis of the grounded linguistic capabilities of fourteen different MLLMs, focusing on ICL and \CoT across the tasks available in the VALSE benchmark.%

\subsection{Chain-of-Thought (\ICL) Prompting}
\label{sec:relatedWorks:CoT}
Recent research shows that models perform better in reasoning, arithmetic, and commonsense tasks when they develop a reasoning process for their answers \citep{wei2022chain}. This method, known as \CoT, was initially introduced for LLMs. The core idea behind \CoT is that by incorporating intermediate reasoning steps enhances the models' reasoning capabilities, leading to improved results. Models effectively utilize \CoT when provided with context, and numerous studies have explored generating context for multimodal tasks to improve both the quality of demonstrations \citep{rubin2021learning, he2023icl} and the reasoning chain \citep{chen2022program, wang2022self}. However, generating detailed, lengthy, and accurate context can be challenging for humans, which is where \MLLMs come into play \citep{wang2024t, zhang2023multimodal}. Additionally, \CoT can be used without context, in a zero-shot manner, where the model is prompted with the phrase, ``\textit{Let's think step by step}'' \citep{kojima2022large}. In multimodal setting, \citet{mitra2024compositional} investigated \CoT, but their analysis involves generating a scene graph from the query image and use this graph in response generation. On the other hand, in our work, we use detailed CoT descriptions of the images in few-shot setting.

\section{Evaluation Strategy} \label{sec:methodology}

In this study, we investigate the zero-shot and few-shot capabilities of \MLLMs through the VALSE benchmark \citep{parcalabescu-etal-2022-valse}. Previous work has separately examined \ICL and \CoT strategies in multimodal contexts \citep{mitra2024compositional, baldassini2024makes, shukor2023beyond}. This study aims to integrate these approaches and provide a comprehensive analysis regarding how the recent MLLMs tackle with visio-linguistic grounding. Below, we begin by providing a brief review of the VALSE benchmark (\S \ref{sec:methodology:valseBenchmark}). We then present the \ICL methodology (\S \ref{sec:methodology:ICLApproach}) employed in our assessment of \MLLMs, explaining our demonstration example selection process. Finally, we discuss the application of the \CoT approach (\S \ref{sec:methodology:CoTApproach}) in our experimental analysis.

\subsection{VALSE Benchmark} \label{sec:methodology:valseBenchmark}
The VALSE \citep{parcalabescu-etal-2022-valse} is a zero-shot foiling benchmark designed to assess the capabilities of  \MLLMs in integrating linguistic constructs with visual contexts. Providing a comprehensive evaluation framework, VALSE encompasses six distinct tasks that thoroughly probe the model’s ability to bridge language and vision. These tasks include \emph{Existence, Plurality, Counting, Spatial Relations, Actions}, and \emph{Coreference,} each focusing on a critical linguistic phenomenon necessary for a deep understanding.

\begin{itemize}[leftmargin=*]
    \item \emph{Existence} task examines the model's ability to identify the presence or absence of entities in an image. Models must differentiate between scenarios where objects exist or not within the visual context, focusing on existential quantifiers.
    \item \emph{Plurality} task tests the model's understanding of singular and plural forms by requiring it to distinguish between images depicting single and multiple instances of objects. It assesses semantic number comprehension.
    \item \emph{Counting} task challenges the model to accurately count the number of entities present in an image. The scenarios vary in complexity, demanding precise enumeration capabilities.
    \item \emph{Spatial Relations} task evaluates the model's ability to recognize and interpret spatial relationships between objects in an image. It focuses on understanding the arrangements and positions of items relative to each other.
    \item \emph{Actions} task assesses the model's proficiency in identifying and understanding actions occurring within images. It requires recognizing the activities depicted and understanding the roles and interactions of the participants involved.
    \item \emph{Coreference} task determines the model's ability to resolve pronoun references within the visual context. It tests whether the \MLLM can correctly link pronouns to the corresponding entities in the images, ensuring coherent understanding.
\end{itemize}

Additionally, VALSE presents foils for \emph{Foil-It!} \cite{shekhar2017foil} dataset which connects objects in the captions to the MS-COCO \cite{lin2014microsoft} dataset. Refer to Appendix \ref{sec:appendix:valseDataset} for further details about VALSE benchmark.

In this work, we aim to investigate the performance of MLLMs on the VALSE benchmark and analyze how few-shot settings can enhance their capabilities in grounding language within visual contexts. Specifically, we focus on models pretrained on interleaved image-text data, which support few-shot learning, to understand the impact of this training strategy. Additionally, we analyze the performance of MLLMs pretrained solely on image captioning data, which do not support few-shot learning, to provide a comprehensive evaluation across different pretraining schemes.

\subsection{Few-Shot \ICL Strategy} \label{sec:methodology:ICLApproach}
Few-shot \ICL aims to increase model performance by providing a few demonstration examples that are contextually related to the query image-text pair. The optimal selection and arrangement~of~these examples is an active area of research \citep{an-etal-2023-context, liu-etal-2022-makes, lu-etal-2022-fantastically, yoo-etal-2022-ground, min-etal-2022-rethinking, chen2023understanding}. Our investigation examines the impact of in-context demonstrations on model performance by comparing randomly selected examples with those closely matching the visual and textual content of the~query~pair.

\noindent \textbf{Example Selection.} For example selection, we employed the~Mixed Modality In-Context Example Selection (MMICES) method \citep{chen2023understanding}. This method assesses both textual and visual cosine similarity between the image-text pairs in the demonstration examples and the query pair. Using CLIP %
as our encoder, we first identified the top $K$ visually similar examples. From these $K$ visually similar examples, we refined the selection to $N$ examples exhibiting textual similarity. The value of $N$ denotes the shot count used in our experiments. 

Determining the appropriate value of $K$ proved to be critical and challenging, as it directly influences the model's exposure to textually similar examples. Our analysis revealed that higher $K$ values yielded improved results. Consequently, we set $K$ to a high value of 100 for our experiments, ensuring that the model received suitable contextual information for learning and enhancement.

\subsection{\CoT Strategy} \label{sec:methodology:CoTApproach}
\CoT approach aims to enhance model performance by promoting reasoning during inference, particularly in scenarios with limited data. Initially, we experimented with zero-shot \CoT, where the model is asked to generate reasoning without providing additional context. However, we found that without this context, models often generate final answers without engaging in any reasoning process. To address this, we included reasoning information with the demonstration examples.

Given that samples in VALSE lack detailed, fine-grained descriptions for image-text pairs, we employed LLaVA-NeXT \citep{liu2024visual} to generate \CoT descriptions for the context demonstrations. Although this model is capable of generating dense captions, it occasionally fabricates incorrect information and hallucinates details. To mitigate these issues, we adopted a prompt proposed by \citet{nori2023can}, instructing the model to generate both reasoning and answers, along with a label-validation step to reduce hallucinations. Despite these measures, some instances still lacked detailed \CoT descriptions even when the answers were correct. Hence, we manually discarded instances with incorrect answers or inadequate \CoT descriptions. We used only the remaining examples in our few-shot \ICL with \CoT experiments, as they provide detailed and contextually rich demonstrations. Details of this process are provided in the Appendix.

\section{Experiments} \label{sec:experiments}
\subsection{Models} \label{sec:experiments:pretrainedILMs}
We evaluated fourteen state-of-the-art MLLMs, each varying in model size and trained on distinct pretraining datasets. Ten of these models were trained on interleaved image-text data, facilitating to run in few-shot scenarios: OpenFlamingo \citep{awadalla2023openflamingo}, Idefics \citep{laurenccon2023obelics}, Idefics2 \citep{laurençon2024matters}, xGen-MM \citep{xgen_mm_phi3_mini}, Qwen-VL-Chat \citep{Qwen-VL}, and MMICL \citep{zhao2023mmicl}. The remaining four were trained solely on captioning datasets: LLaVA-NeXT \citep{liu2024llavanext}, PaliGemma \citep{PaliGemma}, Intern-VL-Chat-V1.5 \citep{chen2023internvl}, and InterLM-XComposer2 \citep{dong2024internlm}. Appendix \ref{sec:appendix:pretrainedILMs} describes these models in detail.

\begin{table*}[!t]

    \caption{Accuracy performance of the evaluated MLLMs, varying in model size and pretraining strategies, evaluated with 0-8 shots across three settings: Random (\textbf{R}), Similar (\textbf{S}), and Similar with Chain of Thought (\textbf{S+C}) settings. In the \textbf{R} setting, few-shot demonstrations are randomly selected. In the \textbf{S} setting, few-shot examples are selected based on visual and textual similarity. In the \textbf{S+C} setting, examples are also selected based on visual and textual similarity but additionally include a \CoT description. Models with the suffix 'I' indicate instruction-tuned versions.}
    \label{tab:all_results}
    \centering
    \renewcommand{\arraystretch}{1.15}
    \resizebox{\linewidth}{!}{
    \begin{threeparttable}
    \begin{tabular}{@{}l@{\;}c@{$\;$}c@{$\;$}c@{$\;\;\;$}c@{$\;$}c@{$\;$}c@{$\;\;\;$}c@{$\;$}c@{$\;$}c@{$\;\;\;$}c@{$\;$}c@{$\;$}c@{$\;\;\;$}c@{$\;$}c@{$\;$}c@{$\;\;\;$}c@{$\;$}c@{$\;$}c@{$\;\;\;$}c@{$\;$}c@{$\;$}c@{$\;\;\;$}c@{$\;$}c@{$\;$}c@{}}
    \toprule
    \multicolumn{25}{c}{\emph{Zero-Shot Setting}} \\
\midrule
  \textbf{Model} & \multicolumn{3}{@{}c@{}}{\textbf{{Existence}}} & \multicolumn{3}{@{}c@{}}{\textbf{{Plurality}}} & \multicolumn{3}{@{}c@{}}{\textbf{{Counting}}} & \multicolumn{3}{@{}c@{}}{\textbf{{Relations}}} & \multicolumn{3}{@{}c@{}}{\textbf{{Action}}} & \multicolumn{3}{@{}c@{}}{\textbf{{Coreference}}} &  \multicolumn{3}{@{}c@{}}{\textbf{{Foil-It!}}} & \multicolumn{3}{@{}c@{}}{\textbf{{Average}}}\\
\midrule      
       LLaVA-NeXT-34B   & \multicolumn{3}{c}{\textbf{97.0}} & \multicolumn{3}{c}{\underline{71.3}} & \multicolumn{3}{c}{\textbf{82.1}} & \multicolumn{3}{c}{\underline{57.4}} & \multicolumn{3}{c}{\underline{70.9}} & \multicolumn{3}{c}{\textbf{70.4}} & \multicolumn{3}{c}{\textbf{87.6}}  & \multicolumn{3}{c}{\underline{76.7}} \\

       PaliGemma-3B   & \multicolumn{3}{c}{76.6} & \multicolumn{3}{c}{63.7} & \multicolumn{3}{c}{74.1} & \multicolumn{3}{c}{47.1} & \multicolumn{3}{c}{64.2} & \multicolumn{3}{c}{51.2} & \multicolumn{3}{c}{81.2}  & \multicolumn{3}{c}{65.4} \\

       Intern-VL-Chat-V1-5-26B  & \multicolumn{3}{c}{\underline{96.2}} & \multicolumn{3}{c}{\textbf{76.5}} & \multicolumn{3}{c}{76.9} & \multicolumn{3}{c}{\textbf{61.3}} & \multicolumn{3}{c}{\textbf{74.2}} & \multicolumn{3}{c}{\underline{69.5}} & \multicolumn{3}{c}{\underline{87.1}}  & \multicolumn{3}{c}{\textbf{77.4}} \\

       InternLM-XComposer2-7B   & \multicolumn{3}{c}{83.0} & \multicolumn{3}{c}{66.5} & \multicolumn{3}{c}{73.7} & \multicolumn{3}{c}{52.5} & \multicolumn{3}{c}{68.8} & \multicolumn{3}{c}{62.2} & \multicolumn{3}{c}{82.0}  & \multicolumn{3}{c}{69.8} \\
    OpenFlamingo-3B    &\multicolumn{3}{c}{36.4} & \multicolumn{3}{c}{9.4} & \multicolumn{3}{c}{14.2} & \multicolumn{3}{c}{9.0} & \multicolumn{3}{c}{8.5} & \multicolumn{3}{c}{32.0} & \multicolumn{3}{c}{11.0} & \multicolumn{3}{c}{17.2}  \\
    OpenFlamingo-3B I    & \multicolumn{3}{c}{48.3} & \multicolumn{3}{c}{48.3} & \multicolumn{3}{c}{45.6} & \multicolumn{3}{c}{44.1} & \multicolumn{3}{c}{46.0} & \multicolumn{3}{c}{25.0} & \multicolumn{3}{c}{43.3}  & \multicolumn{3}{c}{42.9} \\
    OpenFlamingo-4B    & \multicolumn{3}{c}{46.9} & \multicolumn{3}{c}{54.6} & \multicolumn{3}{c}{49.0} & \multicolumn{3}{c}{47.5} & \multicolumn{3}{c}{51.6} & \multicolumn{3}{c}{49.3} & \multicolumn{3}{c}{49.3}  & \multicolumn{3}{c}{49.7} \\
    OpenFlamingo-4B I   & \multicolumn{3}{c}{48.5} & \multicolumn{3}{c}{54.8} & \multicolumn{3}{c}{50.1} & \multicolumn{3}{c}{47.5} & \multicolumn{3}{c}{51.9} & \multicolumn{3}{c}{46.9} & \multicolumn{3}{c}{49.3}  & \multicolumn{3}{c}{49.9} \\
    Idefics-9B    & \multicolumn{3}{c}{44.2} & \multicolumn{3}{c}{46.2} & \multicolumn{3}{c}{47.1} & \multicolumn{3}{c}{53.8} & \multicolumn{3}{c}{48.2} & \multicolumn{3}{c}{26.3} & \multicolumn{3}{c}{50.4}  & \multicolumn{3}{c}{45.2} \\
    Idefics-9B I    & \multicolumn{3}{c}{58.2} & \multicolumn{3}{c}{54.6} & \multicolumn{3}{c}{50.5} & \multicolumn{3}{c}{49.5} & \multicolumn{3}{c}{58.1} & \multicolumn{3}{c}{54.8} & \multicolumn{3}{c}{56.6}  & \multicolumn{3}{c}{54.6} \\
    Idefics2-8B    & \multicolumn{3}{c}{94.7} & \multicolumn{3}{c}{70.3} & \multicolumn{3}{c}{\underline{79.1}} & \multicolumn{3}{c}{53.6} & \multicolumn{3}{c}{59.8} & \multicolumn{3}{c}{69.1} & \multicolumn{3}{c}{82.1}  & \multicolumn{3}{c}{72.7} \\
    xGen-MM-4.6B   & \multicolumn{3}{c}{37.2} & \multicolumn{3}{c}{34.1} & \multicolumn{3}{c}{37.1} & \multicolumn{3}{c}{39.6} & \multicolumn{3}{c}{36.4} & \multicolumn{3}{c}{37.0} & \multicolumn{3}{c}{40.9}  & \multicolumn{3}{c}{37.5} \\
    Qwen-VL-Chat-9.6B    & \multicolumn{3}{c}{82.6} & \multicolumn{3}{c}{46.3} & \multicolumn{3}{c}{68.3} & \multicolumn{3}{c}{48.0} & \multicolumn{3}{c}{41.1} & \multicolumn{3}{c}{58.7} & \multicolumn{3}{c}{61.9}  & \multicolumn{3}{c}{58.1}\\ 
    MMICL-12.1B    & \multicolumn{3}{c}{65.4} & \multicolumn{3}{c}{57.9} & \multicolumn{3}{c}{53.1} & \multicolumn{3}{c}{57.2} & \multicolumn{3}{c}{59.4} & \multicolumn{3}{c}{61.9} & \multicolumn{3}{c}{59.3}  & \multicolumn{3}{c}{59.2}
    \vspace{0.25cm}\\ %

\midrule
       \multicolumn{25}{c}{\emph{4-Shot Setting}} \\
\midrule
         \textbf{{Model}} & \multicolumn{3}{@{}c@{}}{\textbf{{Existence}}} & \multicolumn{3}{@{}c@{}}{\textbf{{Plurality}}} & \multicolumn{3}{@{}c@{}}{\textbf{{Counting}}} & \multicolumn{3}{@{}c@{}}{\textbf{{Relations}}} & \multicolumn{3}{@{}c@{}}{\textbf{{Action}}} & \multicolumn{3}{@{}c@{}}{\textbf{{Coreference}}} &  \multicolumn{3}{@{}c@{}}{\textbf{{Foil-It!}}} & \multicolumn{3}{@{}c@{}}{\textbf{{Average}}}\\
\midrule
    & {R} & {S} & {S+C} & {R} & {S} & {S+C} & {R} & {S} & {S+C} & {R} & {S} & {S+C} & {R} & {S} & {S+C} & {R} & {S} & {S+C} & {R} & {S} & {S+C} & {R} & {S} & {S+C}\\
\cmidrule{2-25}
    OpenFlamingo-3B & 54.5 & 67.9 & 45.7 & 53.2 & 52.2 & 32.7 & 54.3 & 59.3 & 41.5 & 47.7 & 52.9 & 29.9  & 49.0 & 51.9 & 33.0 & 52.7 & 57.2 & 25.4 & 50.8 & 52.8 & 28.4 & 51.7 & 56.3 & 33.8\\

    OpenFlamingo-3B I & 52.1 & 61.6 & 49.3 & 53.4 & 50.5 & 34.1 & 53.4 & 57.4 & 41.1 & 51.0 & 50.1 & 24.5  & 54.2 & 52.7 & 31.1 & 51.5 & 55.0 & 24.0 & 50.7 & 50.2 & 32.0 & 52.3 & 53.9 & 33.7\\

    OpenFlamingo-4B & 53.7 & 73.1 & 43.6 & 50.9 & 52.3 & 42.5 & 54.6 & 58.4 & 39.9 & 50.1 & 54.6 & 28.8  & 57.8 & 57.5 & 30.6 & 50.5 & 52.9 & 31.3 & 48.4 & 53.8 & 33.2 & 52.3 & 57.5 & 35.7\\

    OpenFlamingo-4B I  & 51.9 & 66.1 & 44.6 & 51.9 & 49.2 & 37.6 & 54.1 & 59.2 & 41.2 & 50.5 & 54.6 & 27.3  & 56.2 & 58.3 & 33.7 & 50.8 & 53.0 & 33.0 & 50.0 & 53.1 & 30.1 & 52.2 & 56.2 & 35.6\\

    Idefics-9B  & 59.2 & 81.0 & \underline{87.3} & 49.8 & 54.8 & \underline{73.6} & 54.7 & 61.2 & \underline{79.4} & 50.6 & 52.1 & \textbf{72.9}  & 56.4 & 60.5 & 74.5 & 51.7 & 53.6 & \textbf{82.8} & 57.0 & 59.8 & 69.6 & 54.2 & 60.4 & \textbf{77.2}\\

    Idefics-9B I  & 74.3 & 88.3 & \textbf{87.5} & 58.8 & 58.0 & 69.0 & 59.2 & 65.0 & 78.3 & 54.8 & 57.2 & \underline{70.5}  & 67.5 & \underline{72.9} & \underline{75.7} & 57.3 & 59.2 & \underline{76.5} & 72.2 & 77.9 & \underline{82.7} & 63.4 & 68.3 & \textbf{77.2}\\

    Idefics2-8B  & \underline{83.2} & \textbf{94.3} & 79.8 & \textbf{70.3} & \textbf{69.7} & \textbf{76.6} & \textbf{73.4} & \textbf{71.4} & \textbf{80.1} & \textbf{61.7} & \textbf{63.2} & 70.1 & 70.3 & 72.6 & \textbf{77.0} & \underline{63.3} & 59.8 & 70.7 & \textbf{82.6} & \textbf{84.9} & \textbf{83.1} & \textbf{72.1} & \textbf{73.7} & 76.8\\

    xGen-MM-4.6B-7B  & 65.2 & 77.0 & 73.9 & 56.8 & 58.8 & 71.0 & 55.6 & 57.3 & 72.0 & 51.6 & 56.3 & 69.7  & 61.2 & 67.0 & 67.4 & 54.6 & 57.9 & 67.3 & 63.3 & 70.7 & 78.3 & 58.3 & 63.6 & 71.4\\

    Qwen-VL-Chat-9.6B  & \textbf{85.2} & \underline{92.7} & 85.7 & \underline{66.4} & \underline{64.4} & 67.5 & \underline{68.9} & \underline{69.8} & 76.7 & \underline{60.8} & 60.2 & 57.0 & \underline{71.4} & 72.5 & 67.0 & \textbf{64.8} & \textbf{62.0} & 72.2 & \underline{79.2} & \underline{80.1} & 65.6 & \underline{71.0} & \underline{71.7} & 70.2 \\
    
    MMICL-12.1B & 56.6 & 70.5 & 37.6 & 54.4 & 54.8 & 16.9 & 50.1 & 55.9 & 32.4 & 57.2 & \underline{60.6} & 25.2 & \textbf{75.2} & \textbf{73.0} & 24.9 & 61.8 & \underline{60.5} & 40.2 & 59.7 & 56.6 & 21.7 & 59.3 & 61.7 & 28.4
    \vspace{0.25cm}\\
\midrule
       \multicolumn{25}{c}{\emph{8-Shot Setting}} \\
\midrule
\textbf{{Model}} & \multicolumn{3}{@{}c@{}}{\textbf{{Existence}}} & \multicolumn{3}{@{}c@{}}{\textbf{{Plurality}}} & \multicolumn{3}{@{}c@{}}{\textbf{{Counting}}} & \multicolumn{3}{@{}c@{}}{\textbf{{Relations}}} & \multicolumn{3}{@{}c@{}}{\textbf{{Action}}} & \multicolumn{3}{@{}c@{}}{\textbf{{Coreference}}} &  \multicolumn{3}{@{}c@{}}{\textbf{{Foil-It!}}} & \multicolumn{3}{@{}c@{}}{\textbf{{Average}}}\\
\midrule
        & {R} & {S} & {S+C} & {R} & {S} & {S+C} & {R} & {S} & {S+C} & {R} & {S} & {S+C} & {R} & {S} & {S+C} & {R} & {S} & {S+C} & {R} & {S} & {S+C} & {R} & {S} & {S+C}\\

\cmidrule{2-25}
    OpenFlamingo-3B  & 51.5 & 72.3 & 58.4 & 51.7 & 51.7 & 38.4 & 53.1 & 58.6 & 47.9 & 50.3 & 49.5 & 38.5  & 51.9 & 56.8 & 36.3 & 52.1 & 56.3 & 31.6 & 53.9 & 50.3 & 32.2 & 52.1 & 56.5 & 40.5\\

    OpenFlamingo-3B I  & 51.7 & 65.3 & 51.3 & 50.3 & 53.1 & 35.4 & 53.3 & 57.4 & 41.6 & 53.6 & 46.9 & 32.2  & 49.7 & 59.7 & 31.8 & 52.5 & 57.2 & 26.1 & 52.5 & 50.8 & 32.3 & 51.9 & 55.8 & 35.8\\

    OpenFlamingo-4B & 52.5 & 74.1 & 72.1 & 52.1 & 55.6 & 58.9 & 56.0 & 63.6 & 57.8 & 52.9 & 55.9 & 52.5  & 59.4 & 59.4 & 41.4 & 49.9 & 54.2 & 39.9 & 52.2 & 56.5 & 55.1 & 53.6 & 59.9 & 54.0\\

    OpenFlamingo-4B I  & 49.9 & 64.4 & 56.4 & 52.1 & 52.6 & 47.6 & 54.4 & 60.8 & 53.9 & 49.7 & 55.1 & 41.7  & 60.1 & 60.7 & 47.5 & 53.4 & 59.3 & 44.4 & 52.4 & 57.8 & 39.6 & 53.1 & 58.7 & 47.3\\

   Idefics-9B  & 57.2 & 84.4 & \textbf{92.1} & 48.4 & 55.6 & \textbf{77.9} & 54.8 & 65.3 & \textbf{86.9} & 53.1 & 56.1 & \textbf{83.6}  & 59.0 & 66.5 & \textbf{78.2} & 53.2 & 58.6 & \underline{70.7} & 58.1 & 60.2 & 75.0 & 54.8 & 63.8 & \textbf{80.6}\\

   Idefics-9B I  & 76.2 & 89.9 & 79.2 & 57.2 & 61.0 & 70.2 & 58.5 & 65.2 & 76.1 & 56.6 & 60.8 & 69.2  & 68.2 & 71.4 & \underline{76.4} & 55.6 & 61.5 & 53.4 & 74.3 & 76.3 & \underline{77.4} & 63.8 & 69.4 & 71.7\\

   Idefics2-8B  & \textbf{88.5} & \underline{94.3} & \underline{86.7} & \textbf{70.5} & \textbf{71.6} & \underline{76.2} & \textbf{74.5} & \textbf{72.1} & \underline{83.0} & \underline{59.6} & 61.1 & \underline{71.6} & \underline{72.0} & 71.3 & 75.7 & 61.0 & \underline{65.4} & 68.3 & \underline{82.6} & \textbf{83.9} & \textbf{81.3} & \textbf{72.7} & \textbf{74.2} & \underline{77.5}\\

   xGen-MM-4.6B-7B  & 65.5 & 86.1 & 69.1 & 56.3 & 61.5 & 61.5 & 55.5 & 61.6 & 65.2 & 54.2 & 57.6 & 67.5  & 65.8 & 71.0 & 62.3 & 56.5 & 54.1 & 61.0 & 64.7 & 70.4 & 73.0 & 59.8 & 66.0 & 65.7\\

   Qwen-VL-Chat-9.6B & \underline{84.2} & \textbf{95.3} & 72.9 & \underline{64.2} & \underline{66.5} & 65.8 & \underline{70.0} & \underline{71.7} & 76.1 & \textbf{60.6} & \underline{61.5} & 63.7 & \underline{72.0} & \underline{71.5} & 72.9 & \underline{62.4} & 63.9 & \textbf{76.1} & \textbf{84.6} & \underline{83.5} & 66.2 & \underline{71.1} & \underline{73.4} & 70.5\\
    
    MMICL-12.1B & 63.6 & 78.6 & 38.6 & 53.5 & 56.4 & 14.3 & 47.7 & 52.2 & 31.9 & 58.9 & \textbf{63.4} & 21.1 & \textbf{75.7} & \textbf{71.6} & 19.6 & \textbf{63.5} & \textbf{65.6} & 37.5 & 61.9 & 66.3 & 20.3 & 60.7 & 64.9 & 26.2\\

    \bottomrule
\end{tabular}
    \end{threeparttable}
}
\end{table*}

\subsection{Evaluation Strategy} \label{sec:experiments:evaluationMetrics}
\citet{shukor2023beyond} evaluates the effectiveness of the ITM (Image-Text Matching) method, initially examined within CREPE \citep{ma2023crepe}, which shares several similarities with VALSE. In this method, a sentence is presented to the model, labeled either as a caption or a foil, and the model is asked to determine if the sentence correctly describes the corresponding image. This allows for the measurement of accuracy, providing a quantitative assessment of the model's ability to link visual and linguistic information accurately. In our work, we assess the performance of \MLLMs using this strategy and report the average accuracies accross both individual tasks and overall performance.
\vspace{-0.2em}

\subsection{Results and Analysis} \label{sec:experiments:results}

\begin{table*}[!t]
    \caption{Accuracy performance of the MLLMs pretrained on interleaved image and text data, varying in model size, in the few-shot ICL setting. Demonstrations are selected based on their similarity to the query. For each setting, ($N$) textual similar examples are chosen from ($K$) visual similar examples. The table shows performance across different ($K$) values, specifically 20, 50, and 100. Models with the suffix 'I' indicate instruction-tuned versions.}
    \label{tab:ablation_k_results}
    \centering
    \renewcommand{\arraystretch}{1.15}
    \resizebox{\linewidth}{!}{
        \begin{threeparttable}
    \begin{tabular}{@{}l@{\;\;}c@{$\;$}c@{$\;$}c@{$\;\;\;$}c@{$\;$}c@{$\;$}c@{$\;\;\;$}c@{$\;$}c@{$\;$}c@{$\;\;\;$}c@{$\;$}c@{$\;$}c@{$\;\;\;$}c@{$\;$}c@{$\;$}c@{$\;\;\;$}c@{$\;$}c@{$\;$}c@{$\;\;\;$}c@{$\;$}c@{$\;$}c@{$\;\;\;$}c@{$\;$}c@{$\;$}c@{}}
    \toprule
    \multicolumn{25}{c}{\emph{Zero-Shot Setting}} \\
\midrule
\textbf{Model} & \multicolumn{3}{@{}c@{}}{\textbf{Existence}} & \multicolumn{3}{@{}c@{}}{\textbf{Plurality}} & \multicolumn{3}{@{}c@{}}{\textbf{Counting}} & \multicolumn{3}{@{}c@{}}{\textbf{Relations}} & \multicolumn{3}{@{}c@{}}{\textbf{Action}} & \multicolumn{3}{@{}c@{}}{\textbf{Coreference}} &  \multicolumn{3}{@{}c@{}}{\textbf{Foil-It!}} &  \multicolumn{3}{@{}c@{}}{\textbf{Average}}\\
\midrule
    OpenFlamingo-3B   &\multicolumn{3}{c}{36.4} & \multicolumn{3}{c}{9.4} & \multicolumn{3}{c}{14.2} & \multicolumn{3}{c}{9.0} & \multicolumn{3}{c}{8.5} & \multicolumn{3}{c}{32.0} & \multicolumn{3}{c}{11.0} & \multicolumn{3}{c}{17.2}  \\
    OpenFlamingo-3B I & \multicolumn{3}{c}{48.3} & \multicolumn{3}{c}{48.3} & \multicolumn{3}{c}{45.6} & \multicolumn{3}{c}{44.1} & \multicolumn{3}{c}{46.0} & \multicolumn{3}{c}{25.0} & \multicolumn{3}{c}{43.3} & \multicolumn{3}{c}{42.9} \\
    OpenFlamingo-4B   & \multicolumn{3}{c}{46.9} & \multicolumn{3}{c}{54.6} & \multicolumn{3}{c}{49.0} & \multicolumn{3}{c}{47.5} & \multicolumn{3}{c}{51.6} & \multicolumn{3}{c}{49.3} & \multicolumn{3}{c}{49.3} & \multicolumn{3}{c}{49.7} \\
    OpenFlamingo-4B I  & \multicolumn{3}{c}{48.5} & \multicolumn{3}{c}{54.8} & \multicolumn{3}{c}{50.1} & \multicolumn{3}{c}{47.5} & \multicolumn{3}{c}{51.9} & \multicolumn{3}{c}{46.9} & \multicolumn{3}{c}{49.3} & \multicolumn{3}{c}{49.9} \\
    Idefics-9B   & \multicolumn{3}{c}{44.2} & \multicolumn{3}{c}{46.2} & \multicolumn{3}{c}{47.1} & \multicolumn{3}{c}{\underline{53.8}} & \multicolumn{3}{c}{48.2} & \multicolumn{3}{c}{26.3} & \multicolumn{3}{c}{50.4} & \multicolumn{3}{c}{45.2} \\
    Idefics-9B I   & \multicolumn{3}{c}{58.2} & \multicolumn{3}{c}{54.6} & \multicolumn{3}{c}{50.5} & \multicolumn{3}{c}{49.5} & \multicolumn{3}{c}{58.1} & \multicolumn{3}{c}{54.8} & \multicolumn{3}{c}{56.6} & \multicolumn{3}{c}{54.6} \\
    Idefics2-8B-8B     & \multicolumn{3}{c}{\textbf{94.7}} & \multicolumn{3}{c}{\textbf{70.3}} & \multicolumn{3}{c}{\textbf{79.1}} & \multicolumn{3}{c}{53.6} & \multicolumn{3}{c}{\textbf{59.8}} & \multicolumn{3}{c}{\textbf{69.1}} & \multicolumn{3}{c}{\textbf{82.1}} & \multicolumn{3}{c}{\textbf{72.7}} \\
    xGen-MM-4.6B    & \multicolumn{3}{c}{37.2} & \multicolumn{3}{c}{34.1} & \multicolumn{3}{c}{37.1} & \multicolumn{3}{c}{39.6} & \multicolumn{3}{c}{36.4} & \multicolumn{3}{c}{37.0} & \multicolumn{3}{c}{40.9} & \multicolumn{3}{c}{37.5} \\
    Qwen-VL-Chat-9.6B     & \multicolumn{3}{c}{\underline{82.6}} & \multicolumn{3}{c}{46.3} & \multicolumn{3}{c}{\underline{68.3}} & \multicolumn{3}{c}{48.0} & \multicolumn{3}{c}{41.1} & \multicolumn{3}{c}{58.7} & \multicolumn{3}{c}{\underline{61.9}} & \multicolumn{3}{c}{58.1}\\
    
    MMICL-12.1B    & \multicolumn{3}{c}{65.4} & \multicolumn{3}{c}{\underline{57.9}} & \multicolumn{3}{c}{53.1} & \multicolumn{3}{c}{\textbf{57.2}} & \multicolumn{3}{c}{\underline{59.4}} & \multicolumn{3}{c}{\underline{61.9}} & \multicolumn{3}{c}{59.3}  & \multicolumn{3}{c}{\underline{59.2}}
    \vspace{0.25cm}\\

\midrule
    \multicolumn{25}{c}{\emph{4-Shot Setting}} \\
\midrule
\textbf{Model} & \multicolumn{3}{@{}c@{}}{\textbf{Existence}} & \multicolumn{3}{@{}c@{}}{\textbf{Plurality}} & \multicolumn{3}{@{}c@{}}{\textbf{Counting}} & \multicolumn{3}{@{}c@{}}{\textbf{Relations}} & \multicolumn{3}{@{}c@{}}{\textbf{Action}} & \multicolumn{3}{@{}c@{}}{\textbf{Coreference}} &  \multicolumn{3}{@{}c@{}}{\textbf{Foil-It!}} &  \multicolumn{3}{@{}c@{}}{\textbf{Average}}\\
\midrule
       & {20} & {50} & {100} & {20} & {50} & {100} & {20} & {50} & {100} & {20} & {50} & {100} & {20} & {50} & {100} & {20} & {50} & {100} & {20} & {50} & {100} & {20} & {50} & {100}\\
          \cmidrule{2-25}

    OpenFlamingo-3B  & 65.0 & 67.7 & 67.9 & 55.5 & 52.4 & 52.2 & 57.5 & 59.3 & 59.3 & 52.5 & 49.4 & 52.9  & 53.9 & 50.9 & 51.9 & 56.0 & 52.3 & 57.2 & 54.2 & 57.0 & 52.8 & 56.4 & 55.6 & 56.3\\
    
    OpenFlamingo-3B I & 53.1 & 58.8 & 61.6 & 53.1 & 49.2 & 50.5 & 60.0 & 58.2 & 57.4 & 53.3 & 50.3 & 50.1  & 53.1 & 54.1 & 52.7 & 55.3 & 53.7 & 55.0 & 50.0 & 52.5 & 50.2 & 54.0 & 53.8 & 53.9\\
    OpenFlamingo-4B  & 63.8 & 69.3 & 73.1 & 53.1 & 49.2 & 52.3 & 57.6 & 58.8 & 58.4 & 52.3 & 53.8 & 54.6  & 54.9 & 54.1 & 57.5 & 51.1 & 51.8 & 52.9 & 52.8 & 55.6 & 53.8 & 55.1 & 56.1 & 57.5\\
    OpenFlamingo-4B I  & 62.4 & 63.8 & 66.1 & 50.3 & 45.6 & 49.2 & 57.8 & 59.6 & 59.2 & 51.0 & 53.3 & 54.6  & 55.3 & 57.2 & 58.3 & 51.4 & 52.2 & 53.0 & 52.9 & 53.7 & 53.1 & 54.4 & 55.1 & 56.2\\
    Idefics-9B  & 76.0 & 79.6 & 81.0 & 57.6 & 57.0 & 54.8 & 58.3 & 59.9 & 61.2 & 57.6 & 52.1 & 52.1  & 61.6 & 62.1 & 60.5 & 53.6 & 53.7 & 53.6 & 58.2 & 60.1 & 59.8 & 60.4 & 60.6 & 60.4\\
    Idefics-9B I  & \underline{86.3} & 86.7 & \underline{88.3} & 58.0 & 56.0 & 58.0 & 61.4 & 63.3 & 65.0 & 59.1 & 57.9 & 57.2  & 71.5 & 71.9 & \underline{72.9} & 58.5 & 55.0 & 59.2 & 76.7 & 79.1 & \underline{77.9} & 67.4 & 67.1 & 68.3\\

    Idefics2-8B  & \textbf{92.7} & \textbf{94.3} & \textbf{94.3} & \textbf{71.2} & \textbf{68.2} & \textbf{69.7} & \textbf{71.7} & \textbf{71.9} & \underline{71.4} & \textbf{63.4} & \textbf{63.0} & \textbf{63.2} & \textbf{72.4} & \underline{73.8} & 72.6 & \underline{62.1} & 58.5 & 59.8 & \textbf{84.7} & \textbf{84.2} & \textbf{84.9} & \textbf{74.0} & \textbf{73.4} & \textbf{73.7}\\
    xGen-MM-4.6B  & 74.7 & 78.8 & 77.0 & 61.3 & 61.0 & 58.8 & 55.5 & 56.1 & 57.3 & 59.8 & 60.6 & 56.3 & 68.3 & 66.9 & 67.0 & 56.6 & 54.2 & 57.9 & 69.0 & 71.6 & 70.7 & 63.6 & 64.2 & 63.6\\
   
    Qwen-VL-Chat-9.6B  & 85.2 & \underline{92.7} & 85.7 & \underline{66.4} & \underline{64.4} & \underline{67.5} & \underline{68.9} & \underline{69.8} & \textbf{76.7} & \underline{60.8} & 60.2 & 57.0 & 71.4 & 72.5 & 67.0 & \textbf{64.8} & \textbf{62.0} & \textbf{72.2} & \underline{79.2} & \underline{80.1} & 65.6 & \underline{71.0} & \underline{71.7} & \underline{70.2} \\
    
   MMICL-12.1B & 65.5 & 70.9 & 70.5 & 52.2 & 50.1 & 54.8 & 52.6 & 53.0 & 55.9 & 59.8 & \underline{60.8} & \underline{60.6} & \underline{72.1} & \textbf{74.8} & \textbf{73.0} & 61.0 & \underline{60.4} & \underline{60.5} & 59.9 & 61.2 & 56.6 & 60.4 & 61.6 & 61.7
    \vspace{0.25cm}\\

\midrule
    \multicolumn{25}{c}{\emph{8-Shot Setting}} \\
\midrule
\textbf{Model} & \multicolumn{3}{@{}c@{}}{\textbf{Existence}} & \multicolumn{3}{@{}c@{}}{\textbf{Plurality}} & \multicolumn{3}{@{}c@{}}{\textbf{Counting}} & \multicolumn{3}{@{}c@{}}{\textbf{Relations}} & \multicolumn{3}{@{}c@{}}{\textbf{Action}} & \multicolumn{3}{@{}c@{}}{\textbf{Coreference}} &  \multicolumn{3}{@{}c@{}}{\textbf{Foil-It!}} &  \multicolumn{3}{@{}c@{}}{\textbf{Average}}\\
\midrule
       & {20} & {50} & {100} & {20} & {50} & {100} & {20} & {50} & {100} & {20} & {50} & {100} & {20} & {50} & {100} & {20} & {50} & {100} & {20} & {50} & {100} & {20} & {50} & {100}\\
          \cmidrule{2-25}
          
    OpenFlamingo-3B  & 65.5 & 66.9 & 72.3 & 51.7 & 52.5 & 51.7 & 56.0 & 60.0 & 58.6 & 47.1 & 52.9 & 49.5  & 56.9 & 56.8 & 56.8 & 53.9 & 58.4 & 56.3 & 52.0 & 51.5 & 50.3 & 54.7 & 57.0 & 56.5\\
    OpenFlamingo-3B I  & 56.4 & 62.2 & 65.3 & 49.0 & 53.4 & 53.1 & 56.6 & 58.3 & 57.4 & 48.8 & 52.1 & 46.9  & 57.7 & 56.8 & 59.7 & 53.9 & 58.6 & 57.2 & 51.5 & 54.5 & 50.8 & 53.4 & 56.6 & 55.8\\

    OpenFlamingo-4B  & 59.8 & 69.5 & 74.1 & 52.5 & 51.7 & 55.6 & 60.7 & 61.5 & 63.6 & 52.3 & 53.1 & 55.9  & 63.0 & 60.8 & 59.4 & 52.8 & 55.6 & 54.2 & 55.6 & 57.4 & 56.5 & 56.7 & 58.5 & 59.9\\

    OpenFlamingo-4B I  & 54.6 & 59.8 & 64.4 & 50.9 & 50.2 & 52.6 & 57.5 & 57.8 & 60.8 & 51.8 & 50.3 & 55.1  & 62.5 & 60.5 & 60.7 & 54.4 & 57.0 & 59.3 & 52.7 & 53.0 & 57.8 & 54.9 & 55.5 & 58.7\\
 
  Idefics-9B  & 73.1 & 79.6 & 84.4 & 53.4 & 57.0 & 55.7 & 60.7 & 66.6 & 65.3 & 54.0 & 56.3 & 56.1  & 65.9 & 64.7 & 66.5 & 54.2 & 57.2 & 58.6 & 58.9 & 61.8 & 60.2 & 60.0 & 63.3 & 63.8\\   

   Idefics-9B I  & 81.6 & 84.8 & 89.9 & 61.1 & 61.2 & 61.0 & 62.2 & 65.9 & 65.2 & 59.4 & 57.4 & 60.8  & 72.2 & 72.0 & 71.4 & 56.4 & 60.5 & 61.5 & 76.7 & 76.0 & 76.3 & 67.1 & 68.3 & 69.4\\

    Idefics2-8B  & \textbf{92.5} & \textbf{93.7} & \underline{94.3} & \textbf{70.9} & \textbf{68.7} & \textbf{71.6} & \textbf{72.2} & \textbf{72.5} & \textbf{72.1} & \underline{63.0} & \textbf{62.1} & 61.1 & 72.7 & 71.6 & 71.3 & \underline{63.0} & 62.7 & \underline{65.4} & \textbf{82.9} & \textbf{84.2} & \textbf{83.9} & \textbf{73.9} & \textbf{73.6} & \textbf{74.2}\\

   xGen-MM-4.6B  & 79.6 & 85.0 & 86.1 & 57.9 & 60.3 & 61.5 & 59.6 & 62.8 & 61.6 & 59.4 & 57.9 & 57.6 & \underline{72.8} & 70.9 & 71.0 & 54.4 & 56.5 & 54.1 & 69.9 & 70.0 & 70.4 & 64.8 & 66.2 & 66.0\\

   Qwen-VL-Chat-9.6B  & \underline{90.7} & \underline{92.3} & \textbf{95.3} & \underline{63.9} & \underline{63.6} & \underline{66.5} & \underline{71.8} & \underline{72.3} & \underline{71.7} & \textbf{63.4} & 59.8 & \underline{61.5} & 72.2 & \underline{73.1} & \underline{71.5} & \textbf{66.4} & \textbf{67.2} & 63.9 & \underline{80.8} & \underline{83.1} & \underline{83.5} & \underline{72.7} & \underline{73.1} & \underline{73.4}\\

   MMICL-12.1B & 74.3 & 77.8 & 78.6 & 55.9 & 55.1 & 56.4 & 49.8 & 51.8 & 52.2 & \underline{63.0} & \underline{61.5} & \textbf{63.4} & \textbf{74.0} & \textbf{73.2} & \textbf{71.6} & 62.4 & \underline{64.6} & \textbf{65.6} & 61.3 & 61.6 & 66.3 & 63.0 & 63.7 & 64.9\\

    \bottomrule
\end{tabular}
    \end{threeparttable}
}
\end{table*}

We show the zero-shot and few-shot capabilities of MLLMs trained on interleaved image-text datasets or captioning datasets in Table \ref{tab:all_results}.%
\vspace{-0.46cm}
\cornersize{.1} 
\begin{center}
\ovalbox{\begin{minipage}{0.97\linewidth}
\emph{Observation 1.} Instruction tuning and \ICL help models follow user instructions.
\end{minipage}}
\end{center}
\noindent Given our questions, we expect the \MLLMs to give a Yes/No response. However, in zero-shot setting, some models struggled in producing outputs containing irrelevant information, leading to notably low scores. Instruction tuning or providing demonstration examples to the models through \ICL often help models in following the expected answer templates. For instance, OpenFlamingo-3B and xGen-MM demonstrate this behavior.%

\vspace{-0.46cm}
\cornersize{.1} 
\begin{center}
\ovalbox{\begin{minipage}{0.97\linewidth}
\emph{Observation 2.} Using similar demonstration examples in \ICL significantly enhances performance compared to random examples.
\end{minipage}}
\end{center}
\noindent Employing demonstration examples in the \ICL setting generally improves overall performance. We observe this behavior consistently across the evaluated MLLMs independent from the model size. Notably, examples similar to query image-text pairs significantly enhance performance compared to random examples. For instance, in the 4-shot setting, OpenFlamingo 3B's performance on \emph{Existence} improves from 54.5\% (Random) to 67.9\% (Similar).%

\vspace{-0.46cm}
\cornersize{.1} 
\begin{center}
\ovalbox{\begin{minipage}{0.97\linewidth}
\emph{Observation 3.} Using more similar demonstration examples generally improves overall performance compared to using random demonstrations. 
\end{minipage}}
\end{center}
\noindent \citet{shukor2023beyond} studied atomic foils with the CREPE benchmark \citep{ma2023crepe}, which is similar to the VALSE benchmark in measuring model performance changes when atomic foils completely alter sentence meanings. They showed that increasing the number of random demonstration examples provides almost no gain in this setup. Our results support this finding and show that increasing the random example count can sometimes even deteriorate performance. However, using a higher number of similar examples helps MLLMs perform better. While more random examples make it difficult to establish a link between the context and query, more similar examples enhance this ability. %

\vspace{-0.46cm}
\cornersize{.1} 
\begin{center}
\ovalbox{\begin{minipage}{0.97\linewidth}
\emph{Observation 4.} The \CoT mechanism diminishes the ability to follow instructions acquired through \ICL in OpenFlamingo variants and MMICL, yet enhances the performance of other models in tasks where they struggle under both zero-shot and \ICL settings.
\end{minipage}}
\end{center}
\noindent CoT descriptions in demonstration examples assist models in reasoning about a given image-text pair, significantly aiding in challenging tasks such as counting, relations, and coreference. For example, in the 4-shot setting for OpenFlamingo 3B, performance on \emph{Relations} improves from 50.1\% (S) to 54.6\% (S+C). However, CoT sometimes causes OpenFlamingo variants and MMICL to ignore the expected answer templates. Although they generate reasoning chains as expected, they fail to provide direct answers to the questions, leading to poor performance. However, for the remaining higher capacity models, CoT generally leads to better performances. %

\vspace{-0.46cm}
\cornersize{.1} 
\begin{center}
\ovalbox{\begin{minipage}{0.97\linewidth}
\emph{Observation 5.} With \ICL and \CoT, lower-capacity models trained on interleaved image-text datasets achieve similar or even better performance than larger-capacity models trained on captioning datasets.
\end{minipage}}
\end{center}
\noindent Except for Idefics2, models trained on interleaved image-text datasets exhibit poor zero-shot performance compared to those trained on captioning data. However, with ICL and CoT, these lower-capacity models achieve similar or even better performance than the larger-capacity models trained on captioning datasets. For example, Idefics-9B obtained 77.2\% accuracy when 4-shot ICL and CoT are applied while Intern-VL-Chat-V1-5-26B achieved 76.7\% overall accuracy. 

\vspace{-0.46cm}
\cornersize{.1} 
\begin{center}
\ovalbox{\begin{minipage}{0.97\linewidth}
\emph{Observation 6.} Models prefer demonstrations that are predominantly textually similar to visual ones, resulting in a slight increase in performance.  
\end{minipage}}
\end{center}
\noindent Table \ref{tab:ablation_k_results} shows the performance changes of models pretrained on interleaved image-text datasets across different $K$ values within the \ICL setting. Increasing the value of $K$ provides a larger pool of visually similar examples. Subsequently, when $N$ examples are selected from this pool based on textual similarity, the final demonstration examples tend to exhibit higher textual similarity to the query image-text pair, albeit potentially lower visual similarity. The results indicate a marginal performance improvement with higher $K$, suggesting that models prefer more textually similar examples.

For additional analyses and qualitative examples of few-shot learning settings, see the Appendix.

\section{Conclusion} \label{sec:conclusion}
This work evaluates MLLMs using the VALSE benchmark to assess the impact of \ICL and \CoT. Our findings show that these strategies significantly enhance model performance, especially in tasks requiring complex reasoning and context understanding. 
We identified specific areas where MLLMs excel and where they struggle, emphasizing the importance of training data composition, pretraining strategies, and effective prompting techniques.

One key insight is that MLLMs trained on captioning datasets perform better in zero-shot settings, while those trained on interleaved image-text data benefit more from few-shot learning. This suggests that targeted pretraining and few-shot strategies are crucial for improving model performance in complex tasks. ICL and CoT prompting enable MLLMs to leverage contextual information and reason through intermediate steps. Future research should optimize these strategies and explore additional methods to enhance model robustness and reasoning capabilities. By refining sophisticated reasoning mechanisms, we can develop MLLMs that are more flexible and effective across a wider range of tasks and settings.

\section{Limitations} \label{sec:limitations}
While the VALSE benchmark provides a comprehensive framework, it may not cover all possible linguistic phenomena or real-world scenarios, potentially limiting the generalizability of the findings to other datasets or applications. Moreover, our study evaluates only fourteen state-of-the-art MLLMs, which, although representative, may not encompass the full spectrum of available models and their respective training datasets. For instance, closed-source proprietary models such as GPT-4o~\citep{GPT_4o}, Gemini 1.5 Pro~\citep{GeminiPro_15}, and Claude 3 Opus~\citep{Claude_3} are intentionally left out due to their restricted access, which limits the ability to conduct comprehensive and reproducible evaluations.

\bibliography{main}

\clearpage
\appendix
\section*{Appendix} \label{sec:appendix}
In the following sections, we provide a comprehensive set of supplementary notes detailing various aspects of our work:
\begin{itemize}[leftmargin=*]
    \item \textbf{Detailed Review of VALSE Benchmark (\S \ref{sec:appendix:valseDataset}):} This section elaborates on the VALSE benchmark, outlining the specific tasks it encompasses.
    \item \textbf{Detailed Review of Evaluated Multimodal LLMs (\S \ref{sec:appendix:pretrainedILMs}):} We offer an in-depth review of all evaluated MLLMs, emphasizing their unique characteristics and capabilities.
    \item \textbf{Demonstrations (\S  \ref{sec:appendix:exampleDetails}):} This section describes our methodology for selecting demonstrations and constructing Chain-of-Thought (CoT) descriptions.
    \item \textbf{Further Analysis (\S \ref{sec:appendix:additionalEvaluation}):} We expand on our key findings, providing additional analyses and insights into individual tasks within the VALSE benchmark.
    \item \textbf{Qualitative Examples (\S \ref{sec:appendix:qualitativeAnalsysis}):} We present qualitative examples that illustrate the few-shot learning settings considered in our study.
\end{itemize}

\section{VALSE Benchmark}\label{sec:appendix:valseDataset}
\begin{figure*}[!t]
    \includegraphics[width=\linewidth]{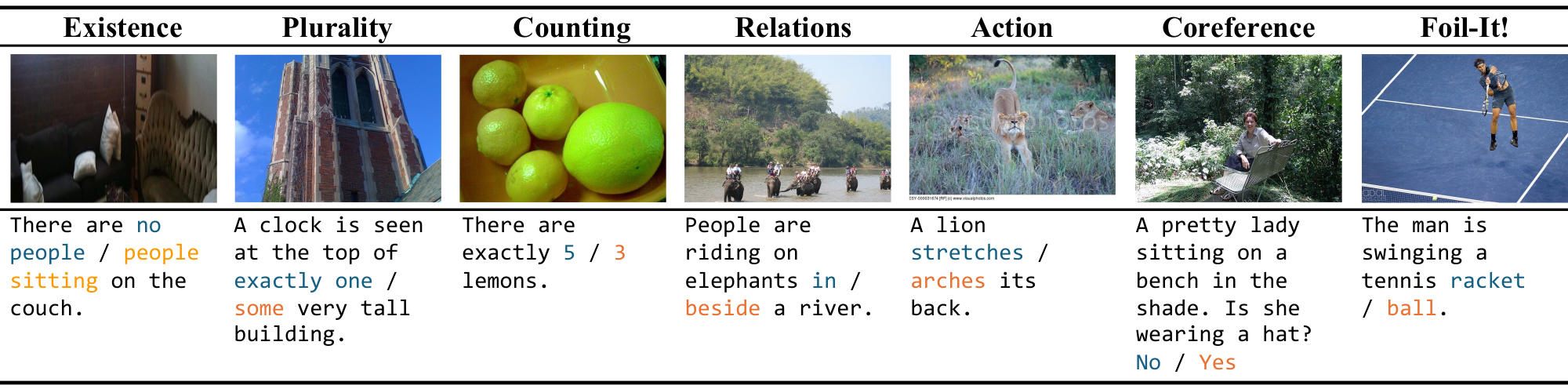}
    \caption{Sample instances from the VALSE benchmark \cite{parcalabescu-etal-2022-valse}.}
    \label{fig:valse-example}
\end{figure*}

The VALSE benchmark \citep{parcalabescu-etal-2022-valse} is a pioneering effort to evaluate the abilities of general-purpose pretrained vision and language models in grounding linguistic constructs within a visual context. It consists of six tasks—\textit{Existence}, \textit{Plurality}, \textit{Counting}, \textit{Spatial Relations}, \textit{Actions}, and \textit{Coreference}—each targeting a key linguistic phenomena  (see Figure \ref{fig:valse-example}). These tasks assess models' capabilities in recognizing existential quantifiers, semantic number, entity counting, spatial arrangements, actions, and pronominal coreference within images, providing a thorough evaluation framework for exploring the complexities of language grounding in visual contexts. The benchmark contains 6795 examples in total.

To develop VALSE, rigorous methodologies were applied to ensure the benchmark's validity and effectiveness \citep{lan2019albert}. This included establishing robust criteria for generating valid foils \citep{xie2019visual}, which are crucial for accurately assessing model performance. Through detailed experimentation and evaluation of five widely-used \MLLMs, the original VALSE paper provided insights into the current challenges faced by pretrained models in understanding and interpreting linguistic phenomena in visual contexts. %

\section{Evaluated \MLLMs}\label{sec:appendix:pretrainedILMs}
\begin{figure*}[!t]
\includegraphics[width=\linewidth]{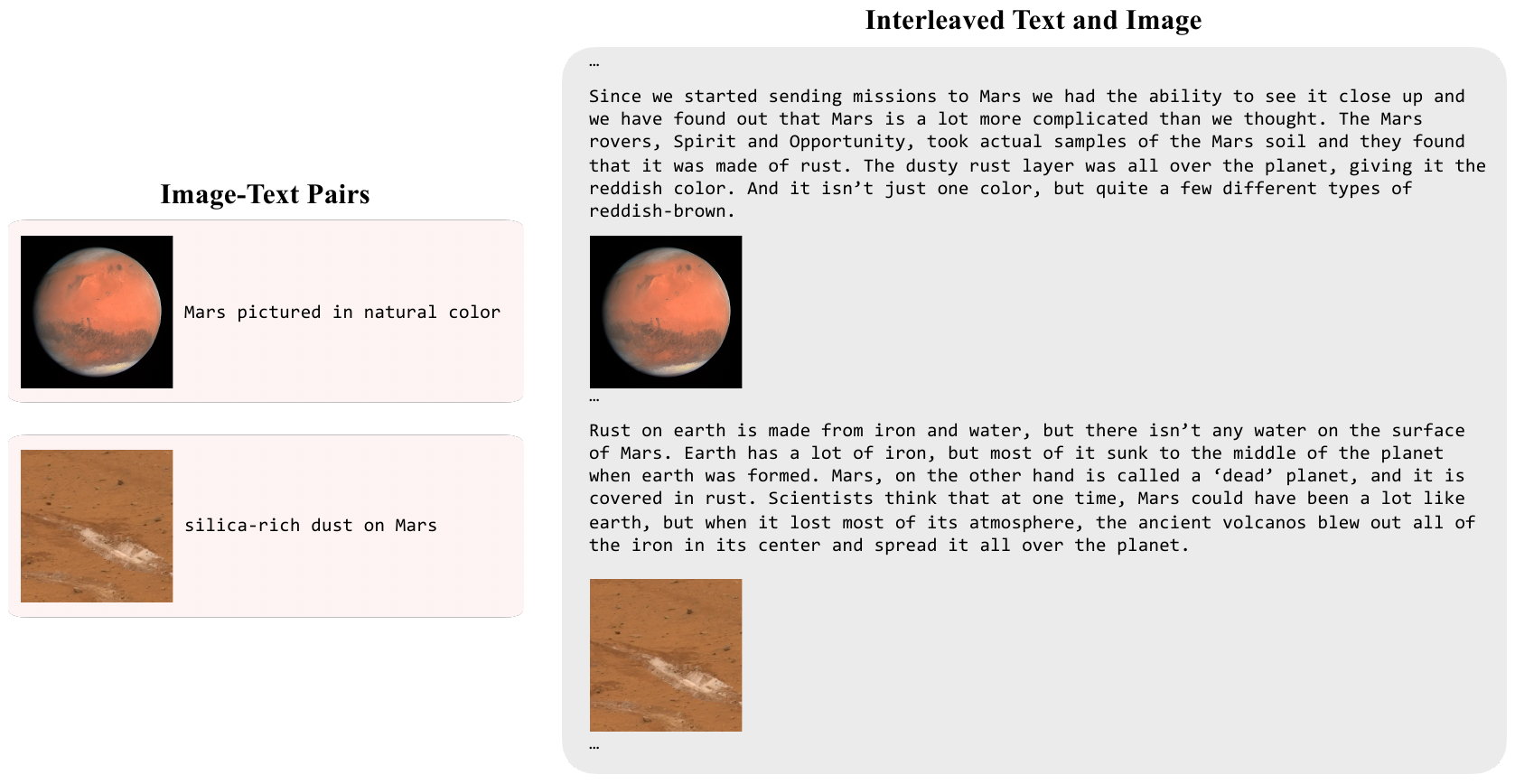}

    \caption{Sample data demonstrating the differences between image-text pairs, and interleaved text and image data used in training \MLLMs.}
    \label{fig:sample-data}
\end{figure*}

Here, we describe the models used in our experiments. We tested models trained on datasets containing image-text pairs (\S\ref{sec:appendix:pretrainedILMsSingle}) as well as models trained on interleaved image-text datasets (\S\ref{sec:appendix:pretrainedILMsMultiple}). Figure~\ref{fig:sample-data} demonstrates sample data that are utilized in each training strategy.

\subsection{\MLLMs pretrained on Captioning Datasets}\label{sec:appendix:pretrainedILMsSingle}
Recently, there has been considerable interest in NLP regarding models capable of handling single image-text pairs \citep{li2023blip2, dai2024instructblip, liu2024visual, zhu2023minigpt, fuyu-8b, ge2023making}. These models demonstrate a remarkable ability to understand and generate textual descriptions for given images, which greatly aids tasks such as image captioning, visual question answering, and image retrieval. By employing sophisticated architectures and multimodal learning techniques, these models effectively integrate visual and textual data to deduce semantic meaning and context. Consequently, they hold significant potential for diverse applications in image comprehension, multimedia analysis, and human-computer interaction.

\noindent \textbf{{LLaVA}} \citep{liu2024visual}, also known as Large Language and Vision Assistant, model family, including LLaVA 1.5 \citep{liu2023improvedllava} and LLaVA-NeXT \citep{liu2024llavanext}, represents a significant leap forward in large multimodal models research. These models surpass natural instruction-following and visual reasoning tasks, with LLaVA 1.5 setting new standards across 12 datasets. The latest iteration, LLaVA-NeXT, enhances reasoning, OCR, and world knowledge capabilities, even outperforming Gemini Pro 1.0 \citep{team2023gemini} on certain benchmarks. LLaVA-NeXT achieves these improvements while maintaining a minimalist design and high data efficiency, requiring fewer than 1M visual instruction tuning samples for training. Notably, it demonstrates leading performance among open-source large multimodal models, %
with significantly lower training costs. During our evaluation, we decided to use the LLaVA-NeXT 34B variant.

\noindent \textbf{{PaliGemma}}, created by Google, is another powerful \MLLM featuring a Transformer decoder and a Vision Transformer image encoder, having 3 billion parameters. Built from Gemma-2B \citep{team2024gemma} and SigLIP-So400m/14 \citep{zhai2023sigmoid}, it follows the PaLI-3 training protocol \citep{chen2023pali}. This model accepts images and text strings as inputs, generating outputs like image captions, answers to questions, object bounding box coordinates, or segmentation codewords. Pre-trained on a variety of datasets including WebLI \citep{chen2023pali}, CC3M-35L \citep{chen2022pali}, VQ²A-CC3M-35L/VQG-CC3M-35L (a subset of VQ2A-CC3M \citep{changpinyo-etal-2022-may}), OpenImages \citep{piergiovanni2022pre}, and WIT \citep{srinivasan2021wit}, PaliGemma surpasses in visual semantic understanding and multilingual tasks. Rigorous data responsibility filters are applied to ensure the training data is safe, clean, and respects privacy by removing inappropriate or sensitive content using advanced filtering techniques.

\noindent \textbf{{Intern-VL-Chat-V1-5}} \citep{chen2024far} is an advanced vision-language model with 26B parameters aimed at closing the performance gap between open-source and commercial models. It utilizes the InternViT-6B \citep{chen2023internvl} vision foundation model and InternLM2-20B \citep{cai2024internlm2} language model, enhanced by three key features: continuous learning with high-quality image-text data, a dynamic high-resolution strategy for detailed image analysis, and a diverse multilingual dataset pipeline. %
In tests across 18 multimodal benchmarks, InternVL 1.5 achieved top results in 8 benchmarks, surpassing leading models like GPT-4V \citep{achiam2023gpt} in OCR-related tasks, showcasing its ability to narrow the gap between open-source and commercial multimodal models. 

\noindent \textbf{{InternLM-XComposer2}} \citep{dong2024internlm}, with 7B parameters, surpasses in generating and comprehending free-form text-image content. By combining text and graphics from diverse inputs such as outlines and reference images, it allows for highly flexible content production beyond traditional comprehension. Utilizing a Partial LoRA (PLoRA) approach to strategically apply additional parameters to image tokens, InternLM-XComposer2 preserves language understanding while enhancing vision comprehension, leading to superior performance in various evaluations compared to existing multimodal models like GPT-4V \citep{achiam2023gpt} and Gemini Pro \citep{team2023gemini}.

\subsection{\MLLMs pretrained on Interleaved Image-Text Data}\label{sec:appendix:pretrainedILMsMultiple}
The development of models capable of handling multiple image-text pairs has become a critical focus in research \cite{awadalla2023openflamingo, laurenccon2023obelics, laurençon2024matters, jiang2024mantis, ye2023mplug, Li2023OtterAM, Qwen-VL, alayrac2022flamingo}. These frameworks demonstrate the ability to analyze and comprehend several instances of image-text pairs simultaneously, enabling a more thorough understanding and interpretation of multimodal data. Through the utilization of advanced multimodal fusion techniques and attention mechanisms, these models seamlessly integrate information from various sources to extract nuanced semantics and context across multiple modalities. This expanded capability broadens the range of applications to tasks such as image album summarization, cross-modal retrieval, and interactive storytelling, where the analysis of multiple image-text pairs enriches the depth and complexity of information processing and comprehension.

\noindent \textbf{{OpenFlamingo}} \citep{awadalla2023openflamingo} introduces a fresh approach to vision and language modeling, enabling autoregressive models to process sequences of mixed images and text for enhanced flexibility, including few-shot learning and multi-round chatbot interactions. Unlike proprietary models such as Flamingo \citep{alayrac2022flamingo}, CM3 \citep{aghajanyan2022cm3}, Kosmos-1 \citep{huang2024language}, PALME \citep{driess2023palm}, and multimodal GPT-4 \citep{achiam2023gpt}, OpenFlamingo provides an open-source alternative, promoting research accessibility. By leveraging pretrained language models with cross-modal attention to vision encoders, OpenFlamingo achieves competitive performance, with models ranging from 3B to 9B parameters. Evaluation across seven datasets indicates that OpenFlamingo models reach 85\% to 89\% of the performance of their corresponding Flamingo models, underscoring their effectiveness and adaptability.

\noindent \textbf{{Idefics}} \citep{laurenccon2023obelics, laurençon2024matters} , includes two versions: Idefics1 and Idefics2. Idefics1, an open-access multimodal model inspired by DeepMind's Flamingo, processes sequences of images and text to generate textual outputs. Utilizing publicly available data and models like CLIP-ViT-H-14 \citep{schuhmann2022laion5b} and LLaMA-65B \citep{touvron2023llama}, it comes in two sizes (80B and 9B parameters) and surpasses image captioning and visual question-answering benchmarks. Idefics2, with 8B parameters, offers improved OCR capabilities, document understanding, and visual reasoning. It handles images in their native resolutions with the NaViT strategy \citep{dehghani2024patch} and incorporates new training data for enhanced OCR and document comprehension.

\noindent \textbf{{xGen-MM}} \citep{xgen_mm_phi3_mini} series, developed by Salesforce AI Research, builds on the successful BLIP series, aligned with Salesforce's XGen initiative for large foundational models. These models, trained on diverse datasets including high-quality image captions, demonstrate state-of-the-art performance in contextual learning. Notably, the xGen-MM mini base model achieves superior performance with under 5 billion parameters, while the fine-tuned xGen-MM mini instruction-tuned model surpasses high-resolution image encoding. Training data sources range from CC12M \citep{changpinyo2021cc12m} to academic VQA tasks, ensuring versatility and robustness. We used the xGen-MM mini base with a model size of 4.6B variant during our experiments. 

\noindent \textbf{{Qwen-VL}} \citep{Qwen-VL} series expands on the Qwen language model, overcoming the limitations of traditional LLMs by integrating visual understanding capabilities. These models, including Qwen-VL-Chat, 9.6B parameters, enable interaction with users through both text and images. They surpass tasks like image captioning and question answering, boasting superior performance and supporting multiple languages. Additionally, Qwen-VL models handle multiple images and demonstrate strong performance across various benchmarks, particularly in fine-grained visual understanding.

\noindent \textbf{{MMICL}} \citep{zhao2023mmicl}, Multi-Modal In-Context Learning, is designed to address the shortcomings of existing \MLLMs in processing complex prompts that involve multiple images and text. MMICL, with a model size of 12.1B, introduces a new method for handling multi-modal inputs, proposes a unique context scheme to improve in-context learning, and utilizes the Multi-modal \ICLLong (MIC) dataset to enhance the model’s ability to understand complex multi-modal prompts. This model effectively tackles challenges such as understanding text-to-image references and the relationships between multiple images. Additionally, MMICL reduces language bias, which often causes \MLLMs to produce hallucinations when dealing with extensive textual contexts.

For our experiments, we follow the model implementations in the HuggingFace repository. We used half-precision to run Idefics1, MMICL, and full precision to run OpenFlamingo variants and xGen-MM. For InterVL-Chat, we applied 8-bit quantization, while the rest of the models were tested with 4-bit quantization. We conducted our experiments on a single Tesla T4, Quadro P4000, V100 or A40 GPU.

\section{Demonstration Examples}\label{sec:appendix:exampleDetails}
\noindent \textbf{Similar Example Selection.} Given the relatively modest size of the VALSE dataset, we opted against partitioning it for creating a demonstration example set. Instead, we leveraged the remaining dataset, excluding the query image-text pair under examination.

\begin{table*}[!t]
    \caption{Rate of valid Chain-of-Thought (CoT) descriptions generated by the corresponding models.}
    \label{tab:cot_generation_results}
    \centering
    \renewcommand{\arraystretch}{1.1}
   \resizebox{\linewidth}{!}{
    \begin{tabular}{lc@{$\;\;$}c@{$\;\;$}c c@{$\;\;$}c@{$\;\;$}c c@{$\;\;$}c@{$\;\;$}c c@{$\;\;$}c@{$\;\;$}c c@{$\;\;$}c@{$\;\;$}c  c@{$\;\;$}c@{$\;\;$}c  c@{$\;\;$}c@{$\;\;$}c }
    \toprule
        \textbf{Model} & {\textbf{{Existence}}} & {\textbf{{Plurals}}} & {\textbf{{Counting}}} & {\textbf{{Relations}}} & {\textbf{{Action}}} & {\textbf{{Coreference}}} &  {\textbf{{Foil-It!}}}\\
    \midrule
    
    LLaVA-NeXT-34B   & 88.3 & 55.2 & 62.4 & 42.2 & 45.8 & 70.9 & 69.8\\
    
    LLaVA-LLAMA3-8B & 5.9 & 20.6 & 6.0 & 17.2 & 15.6 & 16.5 & 7.6\\

    InternLM-XComposer2-7B & 1.8 & 10.3 & 10.8 & 9.7 & 8.3 & 13.8 & 2.3\\

    \bottomrule
    \end{tabular}
    }
    
\end{table*}

\noindent \textbf{\CoTLong Generation.} \CoT approach aims to enhance model performance by promoting reasoning during inference, especially in scenarios with limited data. Initially, we experimented with zero-shot \CoT, where the model generates reasoning without additional context. However, in this setup, models often produced final answers without engaging in reasoning. To address this, we incorporated reasoning information into the demonstration examples. In particular, we employed \MLLMs to generate these \CoT descriptions. The prompt that is used to generate \CoT descriptions is given below:
\cornersize{.1} 
\begin{center}
\vspace{-0.2cm}
\ovalbox{\begin{minipage}{0.97\linewidth}
``{\texttt{\small Given an image and a corresponding sentence, analyze the image to determine if the sentence is true or false. Provide the answer in the format: Final Answer: Yes (if the sentence is true for the image) / No (if the sentence is false for the image). Sentence: ...}}’’
\end{minipage}}
\end{center}
\vspace{0.2cm}

During this process, we encountered challenges such as fabricated information and hallucinated details. To mitigate these issues, we filtered out descriptions yielding incorrect answers. Despite these measures, some instances still lacked \CoT descriptions even when the answers were correct, eventually leading us to discard those with inaccurate or inadequate descriptions and the corresponding samples while selecting the demonstrations for few-shot (ICL + CoT) experiments.

To generate \CoT reasonings and avoid hallucinations, we applied an automatic filtering approach to eliminate some responses. We tested three MLLMs: LLaVA-NeXT 34B \citep{liu2024llavanext}, InternLM-XComposer2 \citep{cai2024internlm2}, and LLaVA-LLaMA3 \citep{2023xtuner}, a LLaVA-1.5-7B \citep{liu2024visual} model finetuned from LLaMA-8B Instruct \citep{llama3modelcard}. Table \ref{tab:cot_generation_results} shows the rate of successful description generation for each model. The results indicate that LLaVA-NeXT clearly surpasses the other models, and larger models generate better reasoning chain descriptions.

\section{Further Analysis}\label{sec:appendix:additionalEvaluation}
In this section, we provide a detailed analysis of the results for each task in VALSE.

\subsection{Existence}
The Existence task is the most basic yet fundamental task in VALSE, assessing a model's ability to determine the presence or absence of an object in an image. All models demonstrated higher accuracy on this task compared to others, indicating that \MLLMs effectively represent objects and determine their existence in a scene. However, when \CoT descriptions were introduced, the performance of all models, except for Idefics-9B, deteriorated. This decline is attributed to the models hallucinating and generating irrelevant reasoning chains in response to the actual question, ultimately leading to incorrect answers. Additionally, as shown in Table \ref{tab:ablation_k_results}, an increase in textually similar examples significantly boosts model performance more than in other tasks.

\subsection{Plurality}

The Plurality task is challenging because the models must not only recognize the given object but also determine its plural form. Results reveal that demonstration examples do not improve the models' understanding of pluralism, although the models correctly recognize the objects. For this task, \CoT reasoning is useful as it directly provides reasoning chains that describe what a plural form is. With this context, models are able to develop an understanding of the task.

\subsection{Counting}
The Counting task, similar to Plurality, evaluates a model's understanding of the exact count of an item in a scene. The model must identify both the object and the number of its appearances. Models trained on captioning datasets outperform those trained on interleaved image-text data. However, the combination of few-shot \ICL and \CoT reasoning enhances the performance of these models, bringing them closer to those trained on captioning data. As seen in qualitative examples, models are guided to count each occurrence, allowing for a direct comparison between the actual and stated occurrences.

\subsection{Spatial Relations}
The Spatial Relations task evaluates models' abilities to recognize interactions between objects. Zero-shot performance shows that all models struggle with this task, as it requires a deep understanding of the interactions and relationships between objects. Results indicate that providing demonstration examples through \ICL helps models achieve a certain performance level, but increasing the number of demonstrations does not lead to further improvement. Performance gains saturate with a higher example count. However, using few-shot \ICL combined with \CoT reasoning, it is possible to achieve up to a 30\% performance increase (Idefics-9B).

\subsection{Action}
The Action task aims to assess how successfully models detect actions and actors in a scene. This task is relatively hard as it requires models to accurately identify dynamic interactions and context-specific activities within an image, which demands a deeper understanding beyond static object recognition. In this task, models trained on captioning data performed better compared to those trained on interleaved image-text datasets. Few-shot \ICL successfully elevated these models' performance to up to 73\%. However, except for the Idefics model family, none of the models benefited from \CoT descriptions. Additionally, increasing the number of demonstration examples did not always positively impact performance.

\subsection{Coreference}
The Coreference task evaluates a model's capability to resolve pronoun references within a visual context, examining whether \MLLMs can accurately associate pronouns with their corresponding entities in images to maintain coherent understanding. This task is challenging as the models need to accurately interpret and maintain contextual relationships between pronouns and their antecedents within a visual scene. Results show that models trained on captioning datasets outperformed those trained on interleaved image-text datasets. The Idefics model family, in particular, substantially benefited from \CoT descriptions, which are crucial for solving the coreference task as they provide explicit reasoning pathways to link pronouns correctly. However, the general model performance with CoT descriptions does not improve much as the number of demonstration examples increases.

\subsection{Foil-It!}
The Foil-It! task is designed to evaluate a model's understanding of objects by replacing the target object with an irrelevant one to create a foil. This task demands models to not only recognize objects accurately but also to detect subtle inconsistencies in the context. Similar to the Counting task, the zero-shot performance of models trained on captioning data surpassed that of models trained on interleaved image-text datasets. Additionally, these models could not be outperformed even with the application of few-shot \ICL and \CoT techniques.

\section{Qualitative Examples}\label{sec:appendix:qualitativeAnalsysis}

\begin{figure*}[!h]
\centering
    \includegraphics[width=\linewidth]{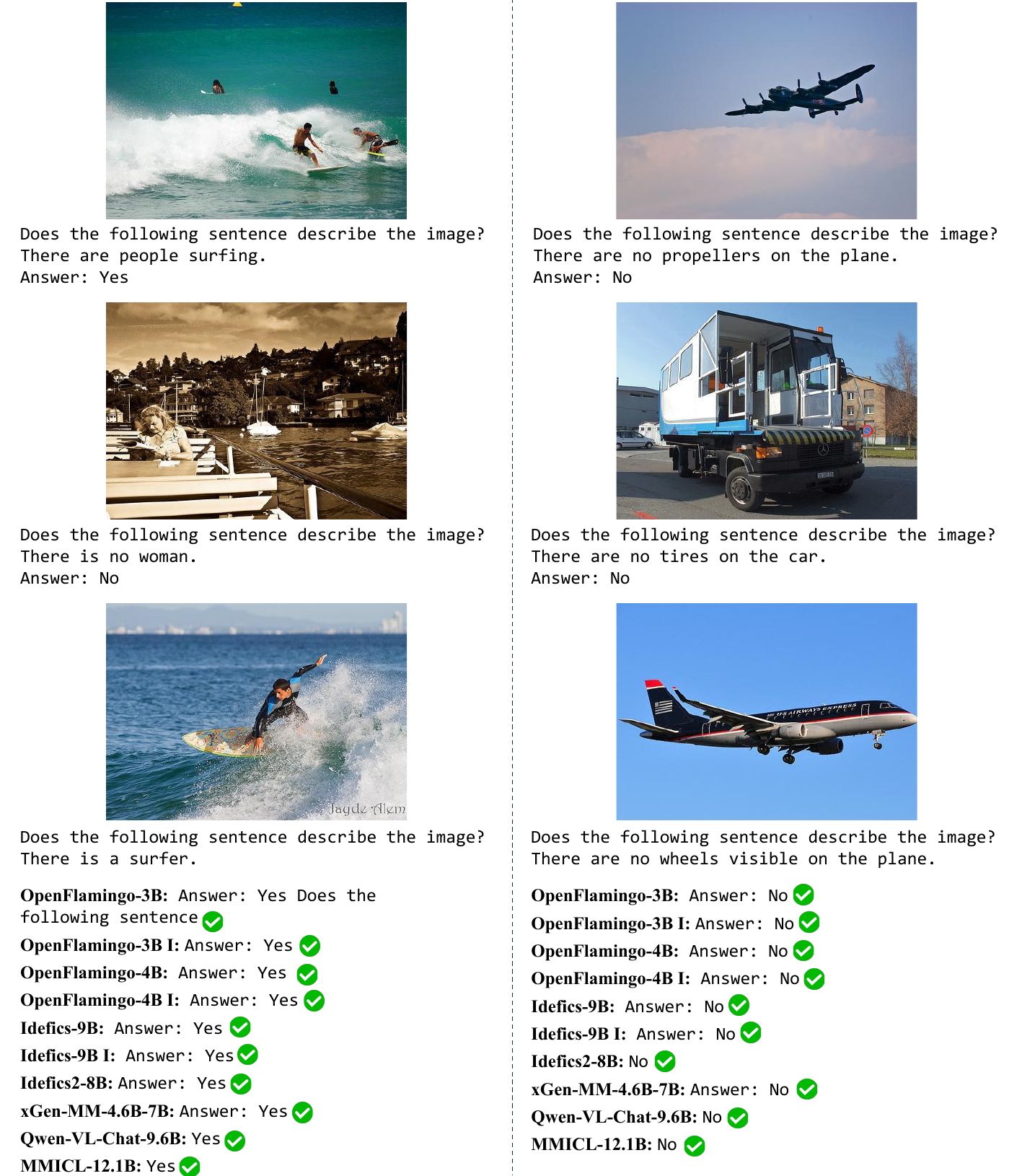}
    \caption{Example model predictions on instances from the \emph{Existence} task, with demonstrations selected based on both visual and textual similarity (setting \textbf{S}).}
    \label{fig:icl-existence-detailed-examples}
\end{figure*}

\begin{figure*}[!t]
\centering
    \includegraphics[width=\linewidth]{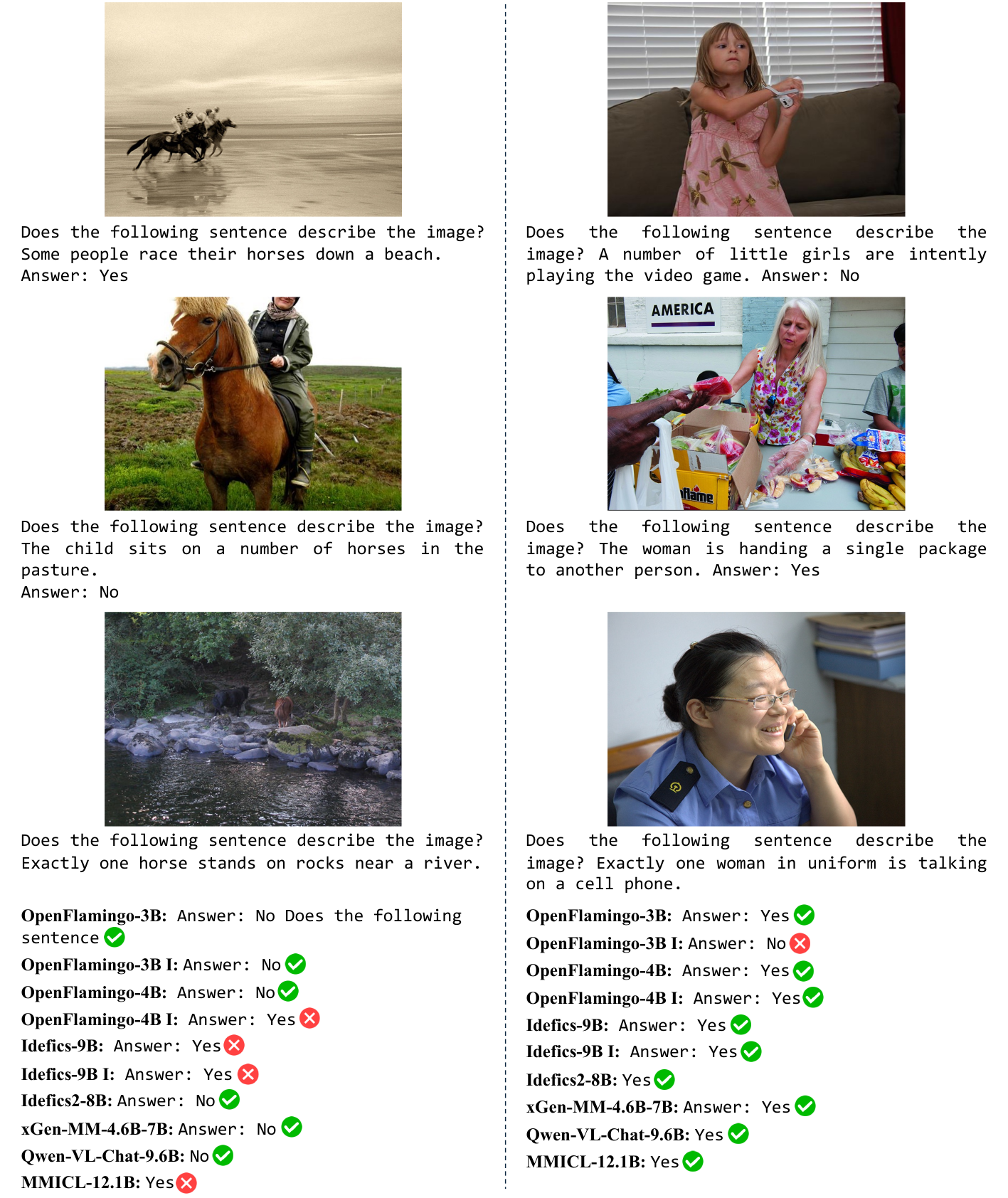}
    \caption{Example model predictions on instances from the \emph{Plurality} task, with demonstrations selected based on both visual and textual similarity (setting \textbf{S}).}
    \label{fig:icl-plurals-detailed-examples}
\end{figure*}

\begin{figure*}[!t]
\centering
    \includegraphics[width=\linewidth]{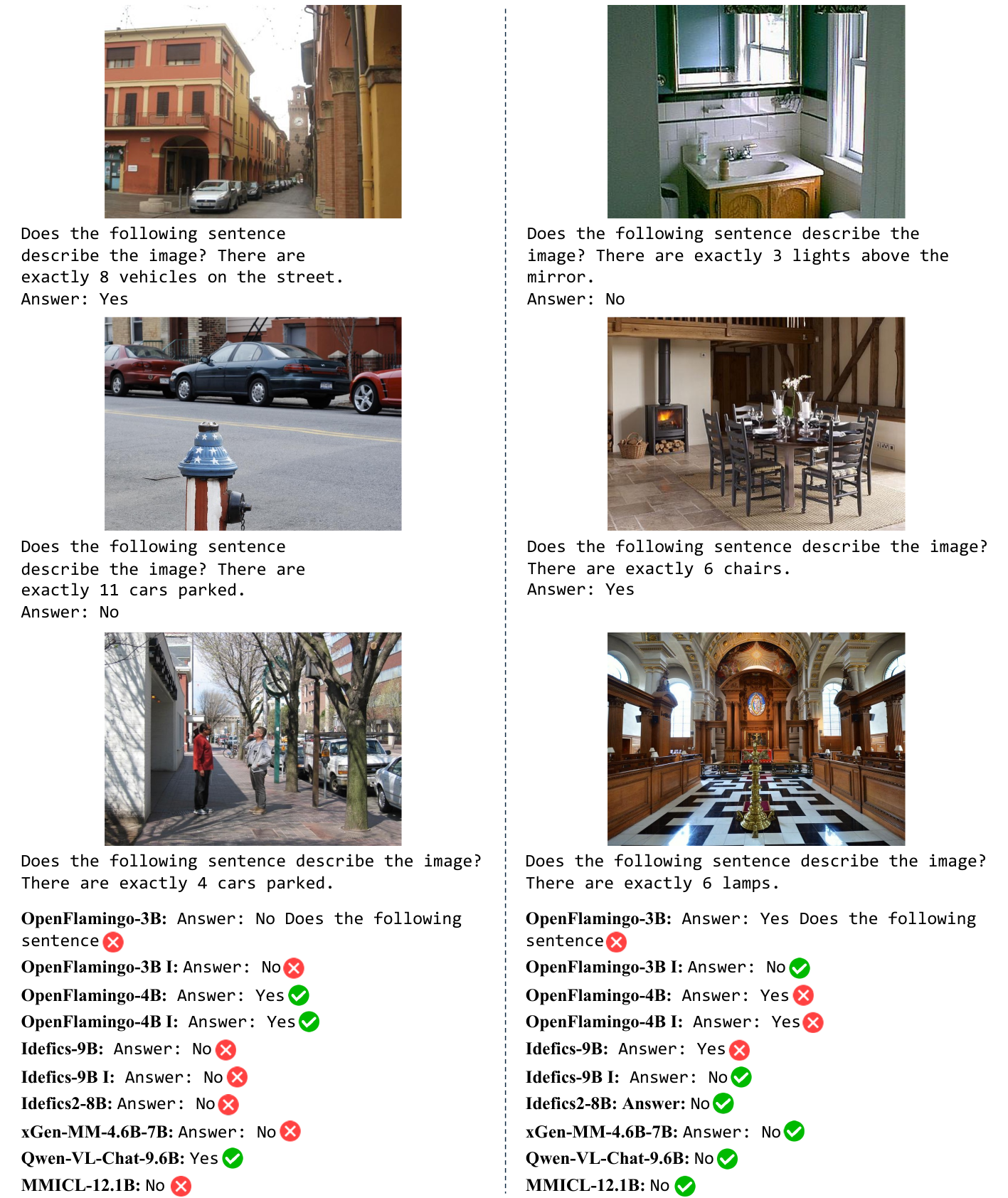}
    \caption{Example model predictions on instances from the \emph{Counting} task, with demonstrations selected based on both visual and textual similarity (setting \textbf{S}).}
    \label{fig:icl-counting-detailed-examples}
\end{figure*}

\begin{figure*}[!t]
\centering
    \includegraphics[width=\linewidth]{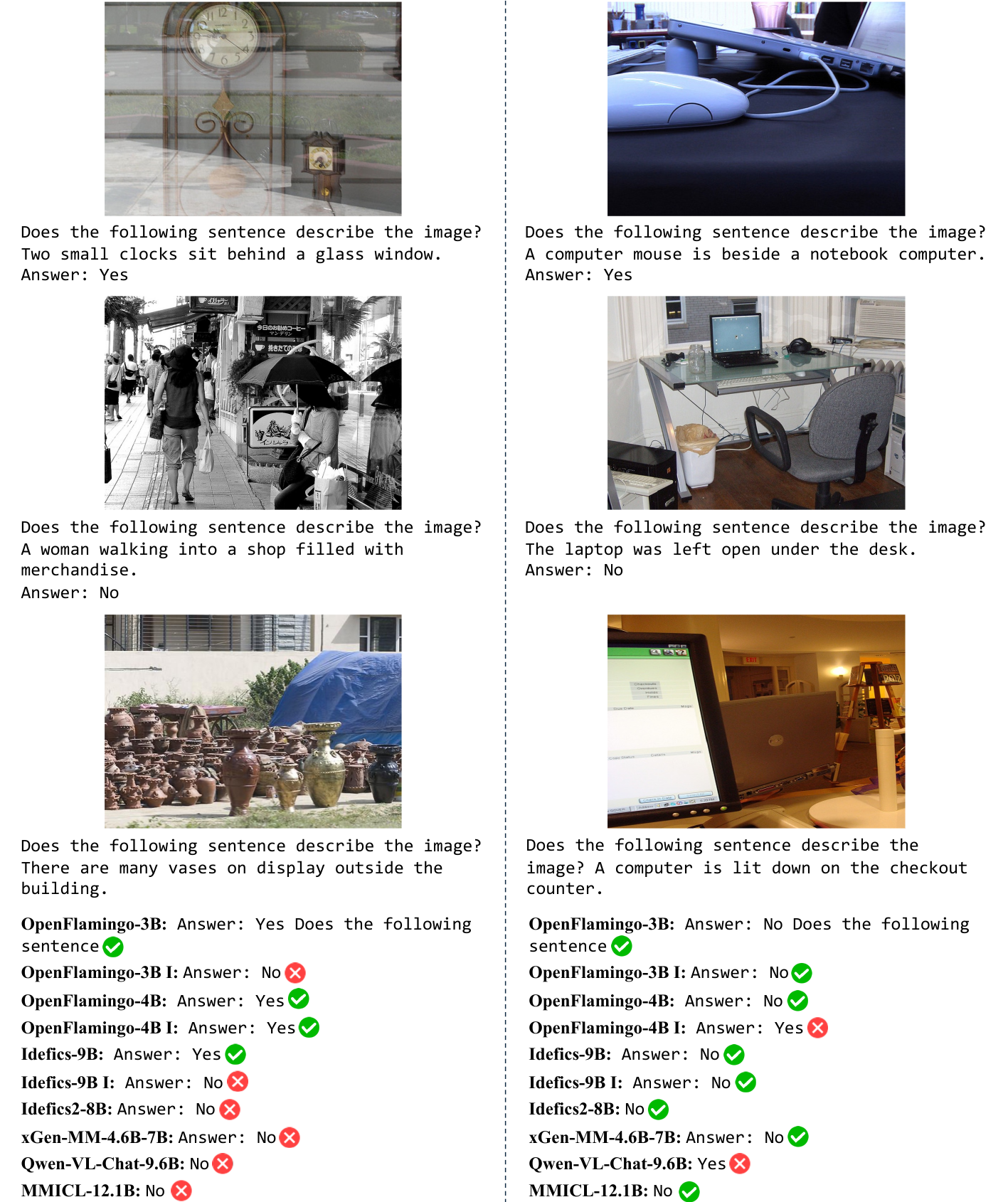}
    \caption{Example model predictions on instances from the \emph{Spatial Relations} task, with demonstrations selected based on both visual and textual similarity (setting \textbf{S}).}
    \label{fig:icl-relations-detailed-examples}
\end{figure*}

\begin{figure*}[!t]
\centering
    \includegraphics[width=\linewidth]{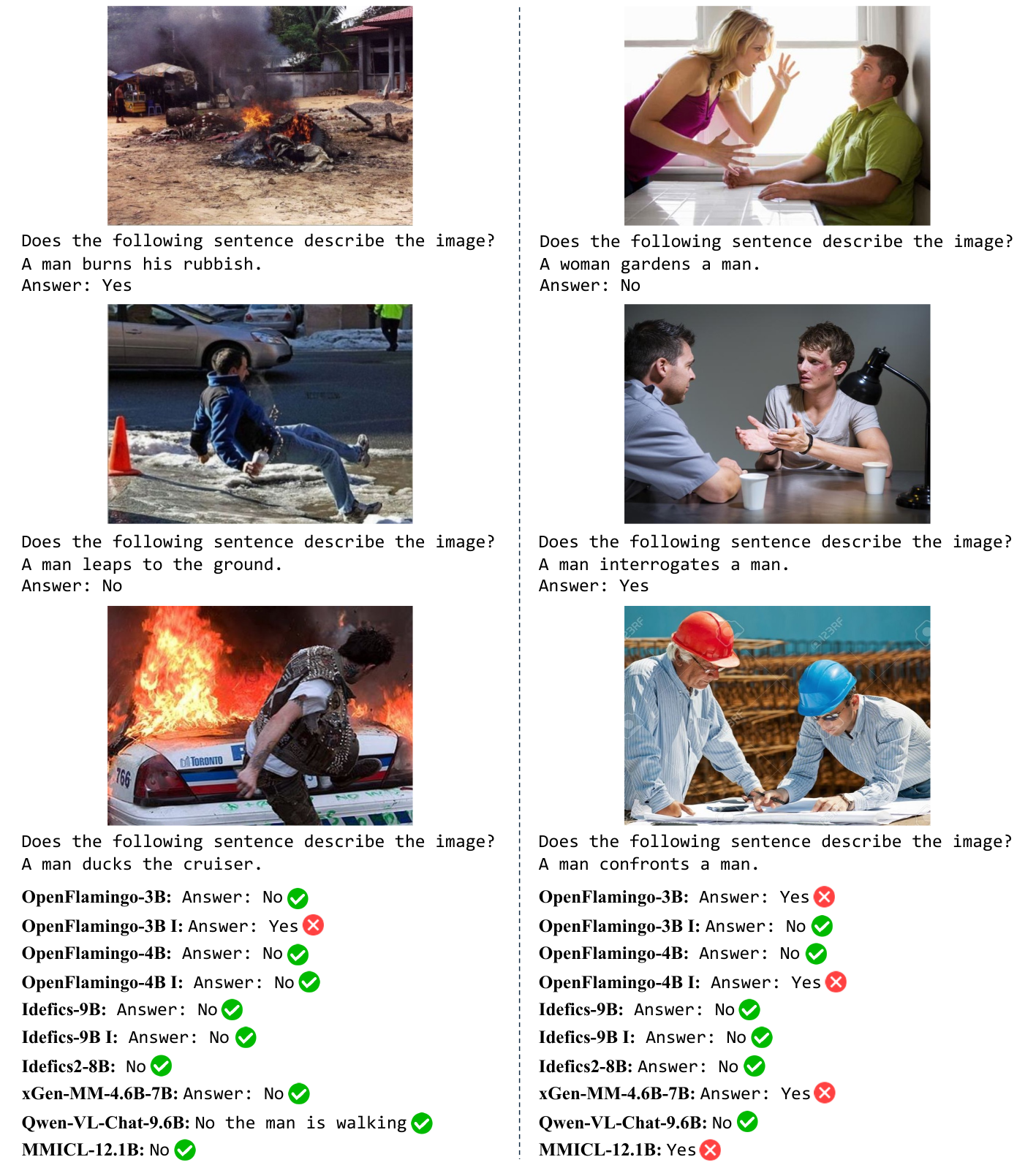}
    \caption{Example model predictions on instances from the \emph{Actions} task, with demonstrations selected based on both visual and textual similarity (setting \textbf{S}).}
    \label{fig:icl-actions-detailed-examples}
\end{figure*}

\begin{figure*}[!t]
\centering
    \includegraphics[width=\linewidth]{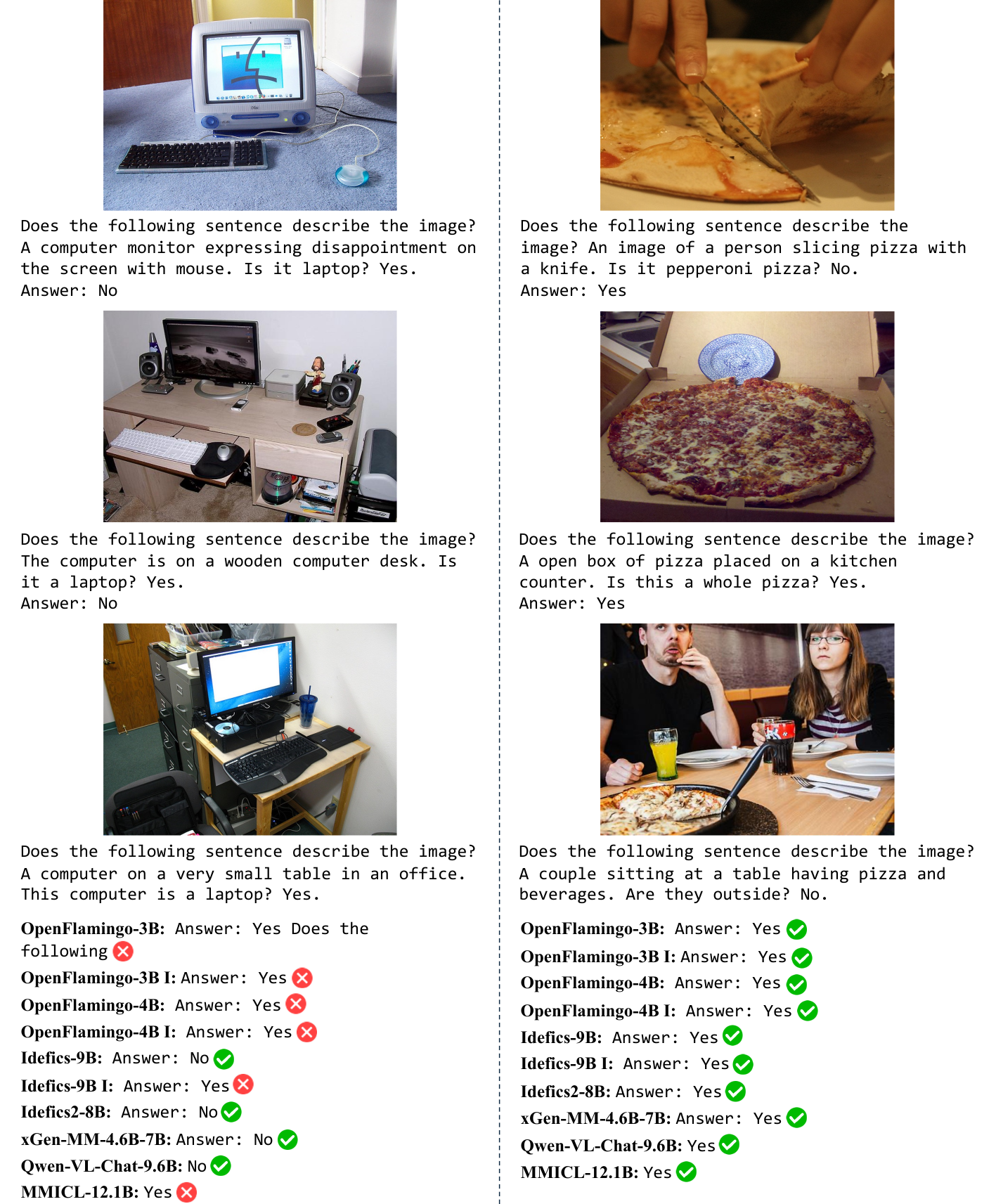}
    \caption{Example model predictions on instances from the \emph{Coreference} task, with demonstrations selected based on both visual and textual similarity (setting \textbf{S}).}
    \label{fig:icl-coreference-detailed-examples}
\end{figure*}

\begin{figure*}[!t]
\centering
    \includegraphics[width=\linewidth]{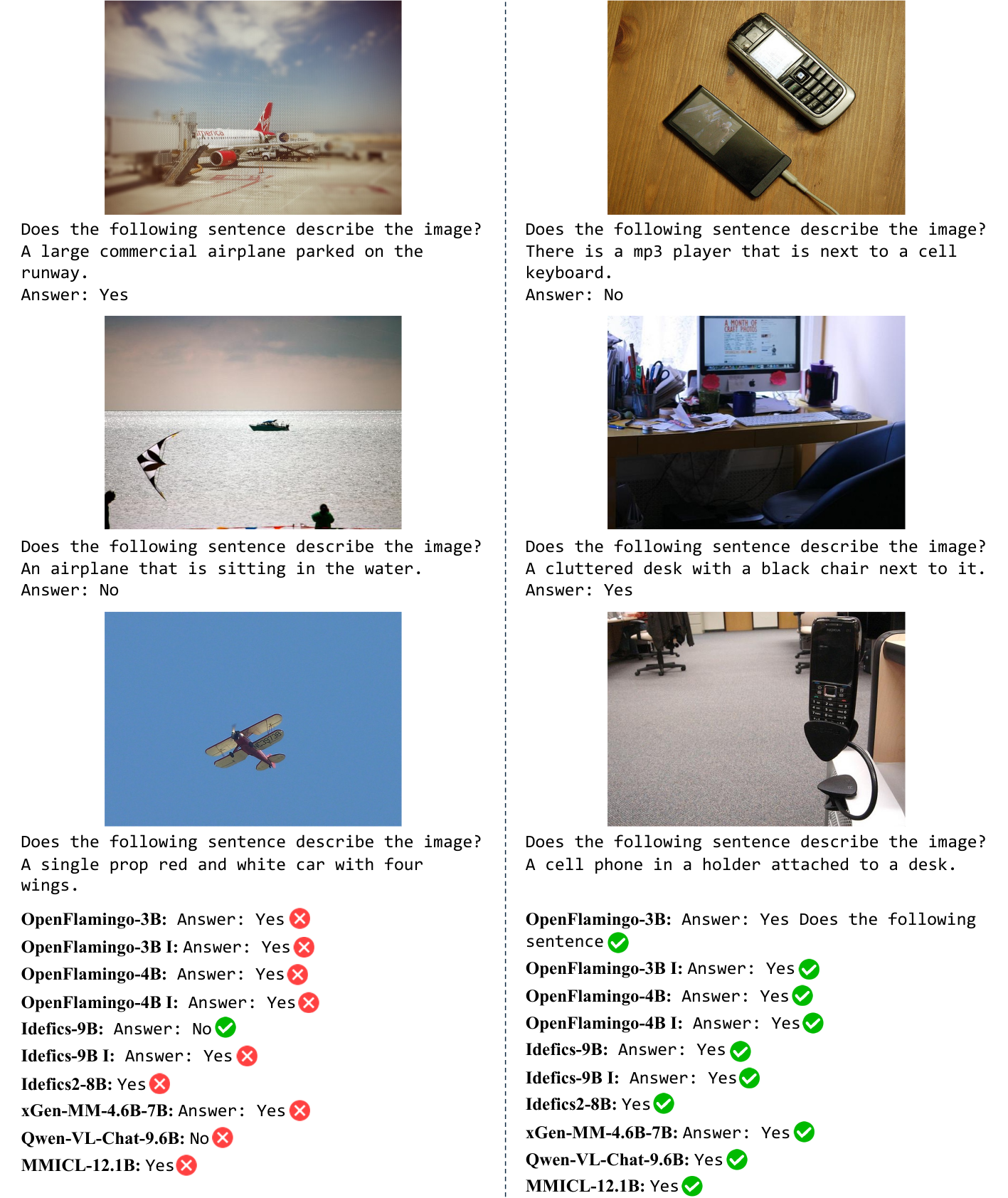}
    \caption{Example model predictions on instances from the \emph{Foil-It!} task, with demonstrations selected based on both visual and textual similarity (setting \textbf{S}).}
    \label{fig:icl-foilit-detailed-examples}
\end{figure*}

\begin{figure*}[!t]
\centering
    \includegraphics[width=\linewidth]{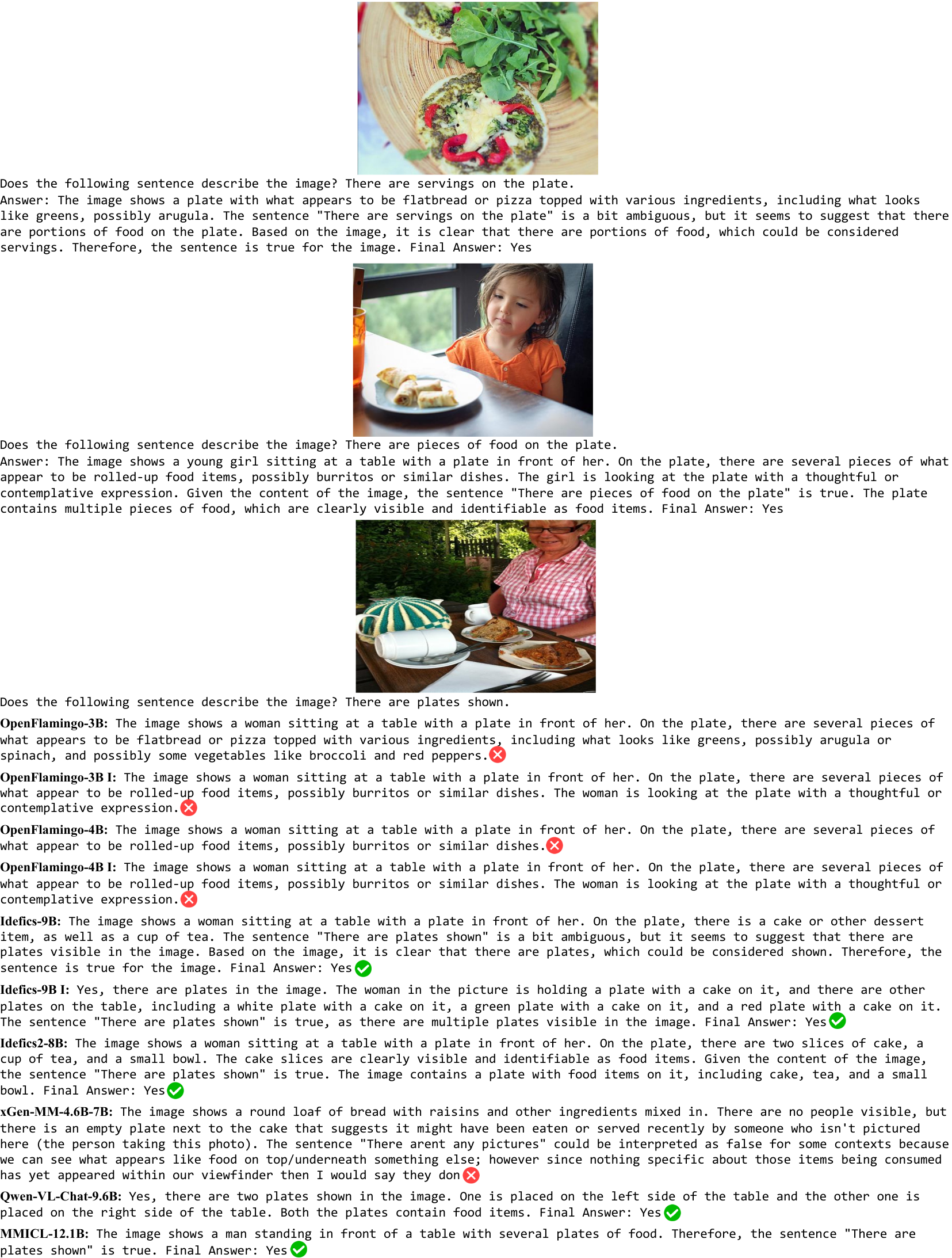}
    \caption{Example model predictions on instances from the \emph{Existence} task, where demonstrations are selected based on visual and textual similarity, and Chain-of-Thought (CoT) reasoning is employed (setting \textbf{S+C}).}
    \label{fig:cot-existence-detailed-examples}
\end{figure*}

\begin{figure*}[!t]
\centering
    \includegraphics[width=\linewidth]{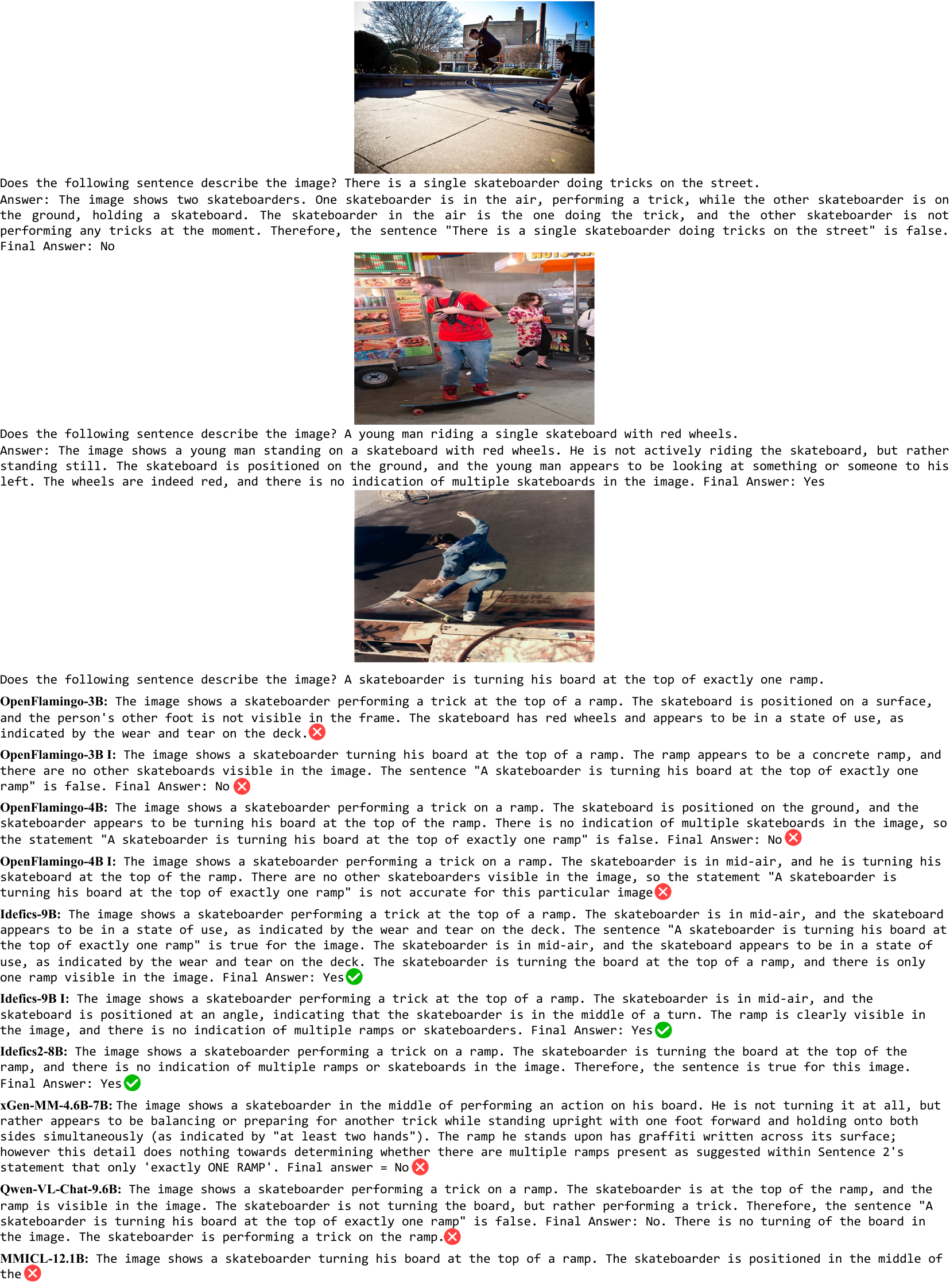}
    \caption{Example model predictions on instances from the \emph{Plurality} task, where demonstrations are selected based on visual and textual similarity, and Chain-of-Thought (CoT) reasoning is employed (setting \textbf{S+C}).}
    \label{fig:cot-plurals-detailed-examples}
\end{figure*}

\begin{figure*}[!t]
\centering
    \includegraphics[width=\linewidth]{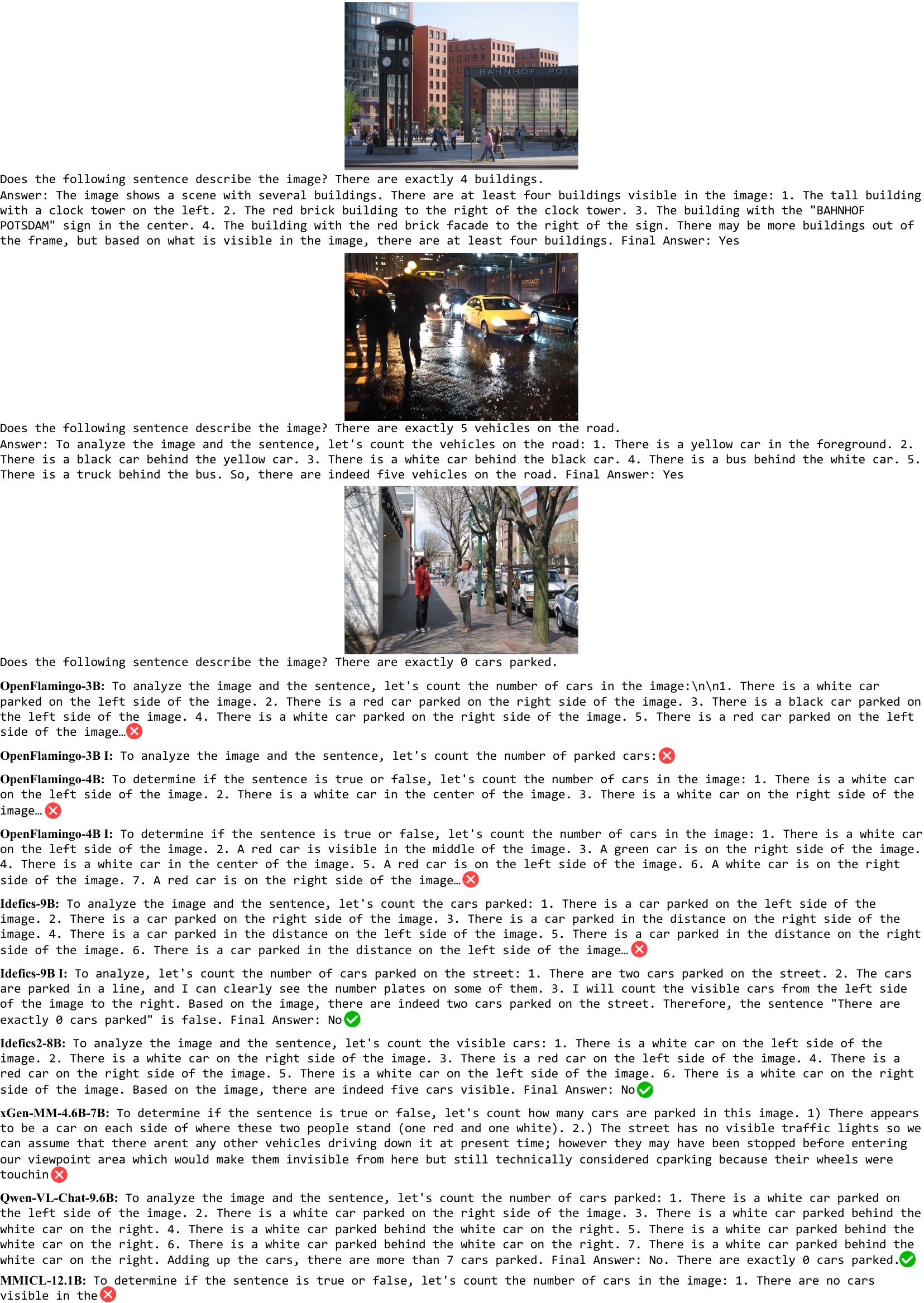}
    \caption{Example model predictions on instances from the \emph{Counting} task, where demonstrations are selected based on visual and textual similarity, and Chain-of-Thought (CoT) reasoning is employed (setting \textbf{S+C}).}
    \label{fig:cot-counting-detailed-examples}
\end{figure*}

\begin{figure*}[!t]
\centering
    \includegraphics[width=\linewidth]{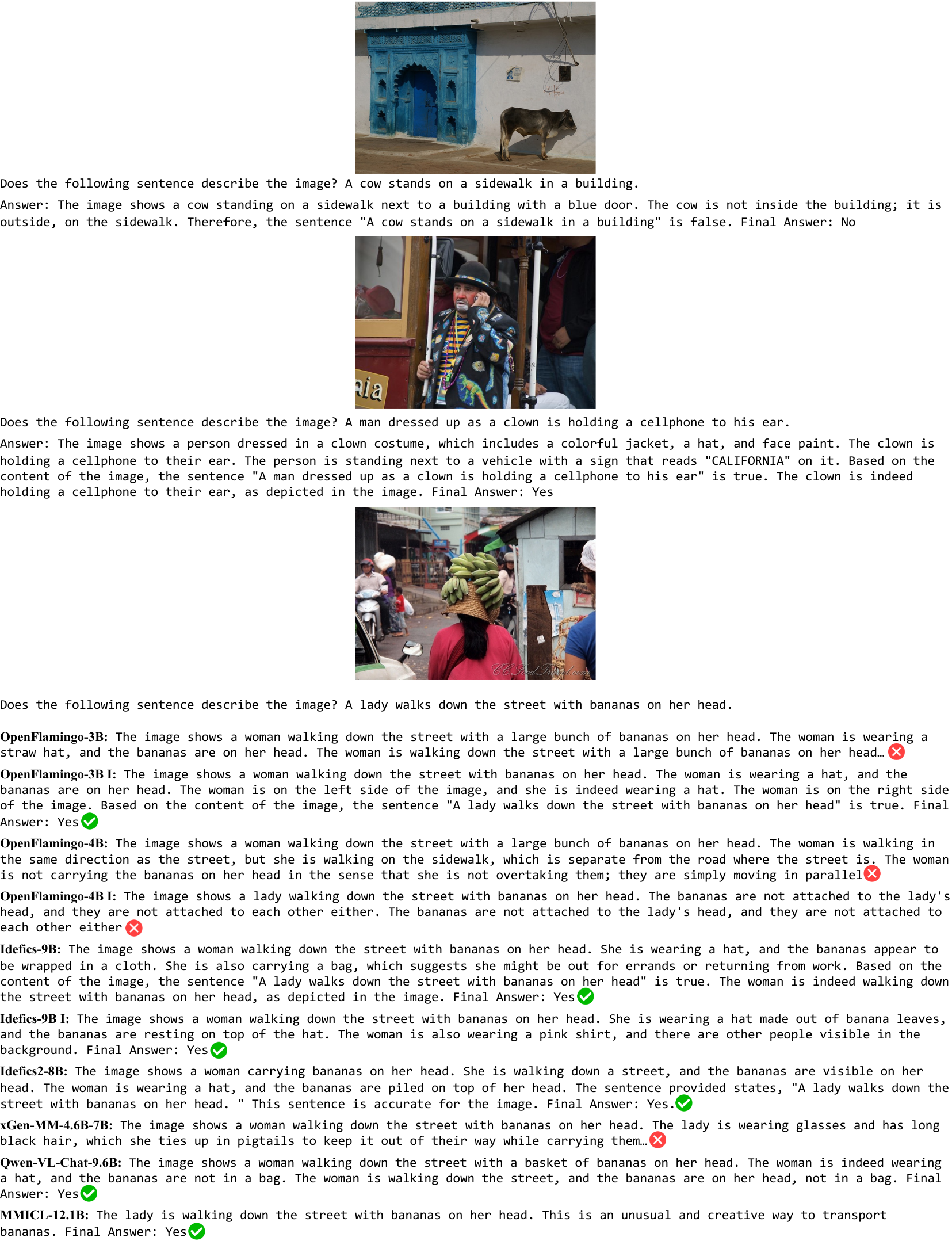}
    \caption{Example model predictions on instances from the \emph{Spatial Relations} task, where demonstrations are selected based on visual and textual similarity, and Chain-of-Thought (CoT) reasoning is employed (setting \textbf{S+C}).}
    \label{fig:cot-relations-detailed-examples}
\end{figure*}

\begin{figure*}[!t]
\centering
    \includegraphics[width=\linewidth]{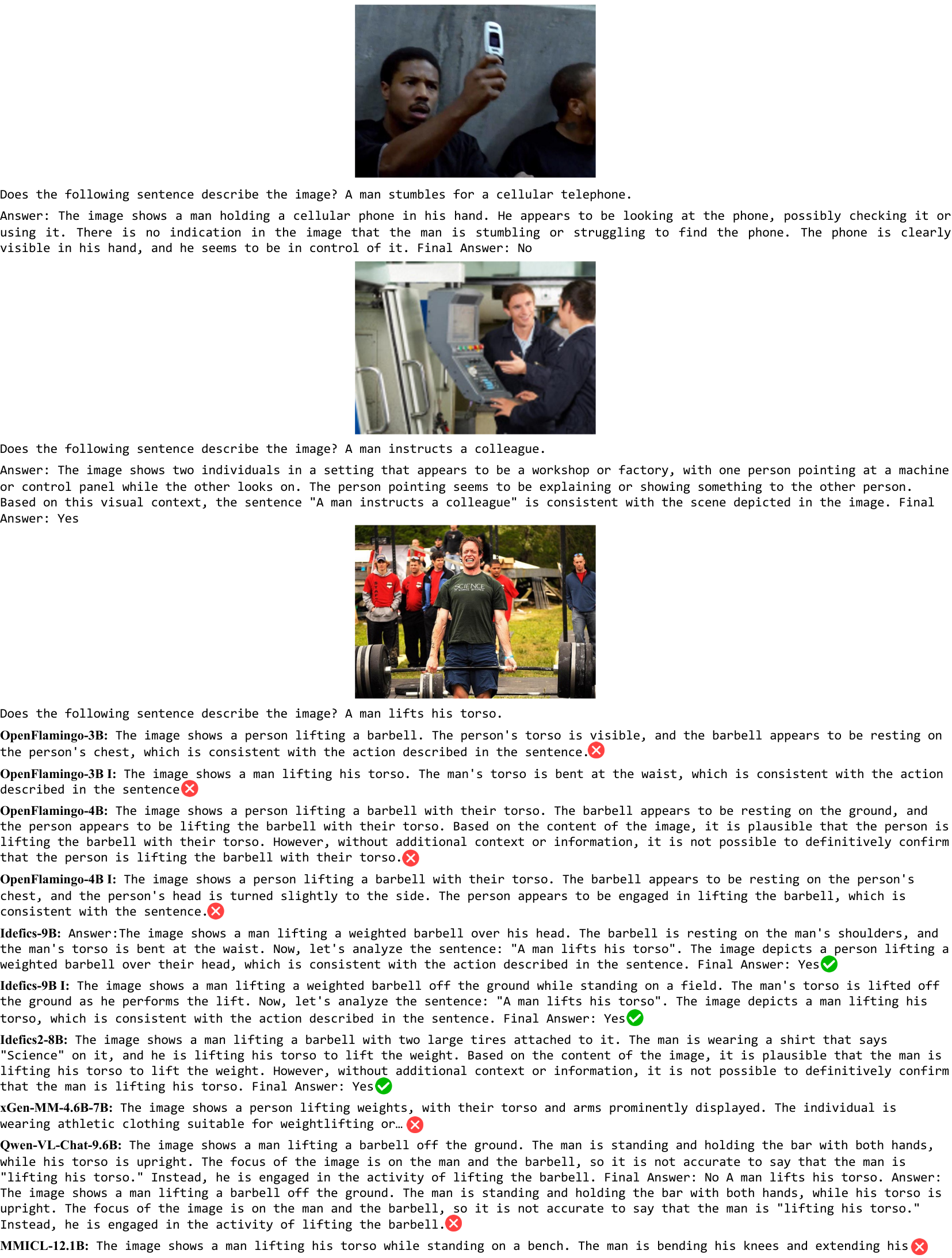}
    \caption{Example model predictions on instances from the \emph{Actions} task, where demonstrations are selected based on visual and textual similarity, and Chain-of-Thought (CoT) reasoning is employed (setting \textbf{S+C}).}
    \label{fig:cot-action-detailed-examples}
\end{figure*}

\begin{figure*}[!t]
\centering
    \includegraphics[width=\linewidth]{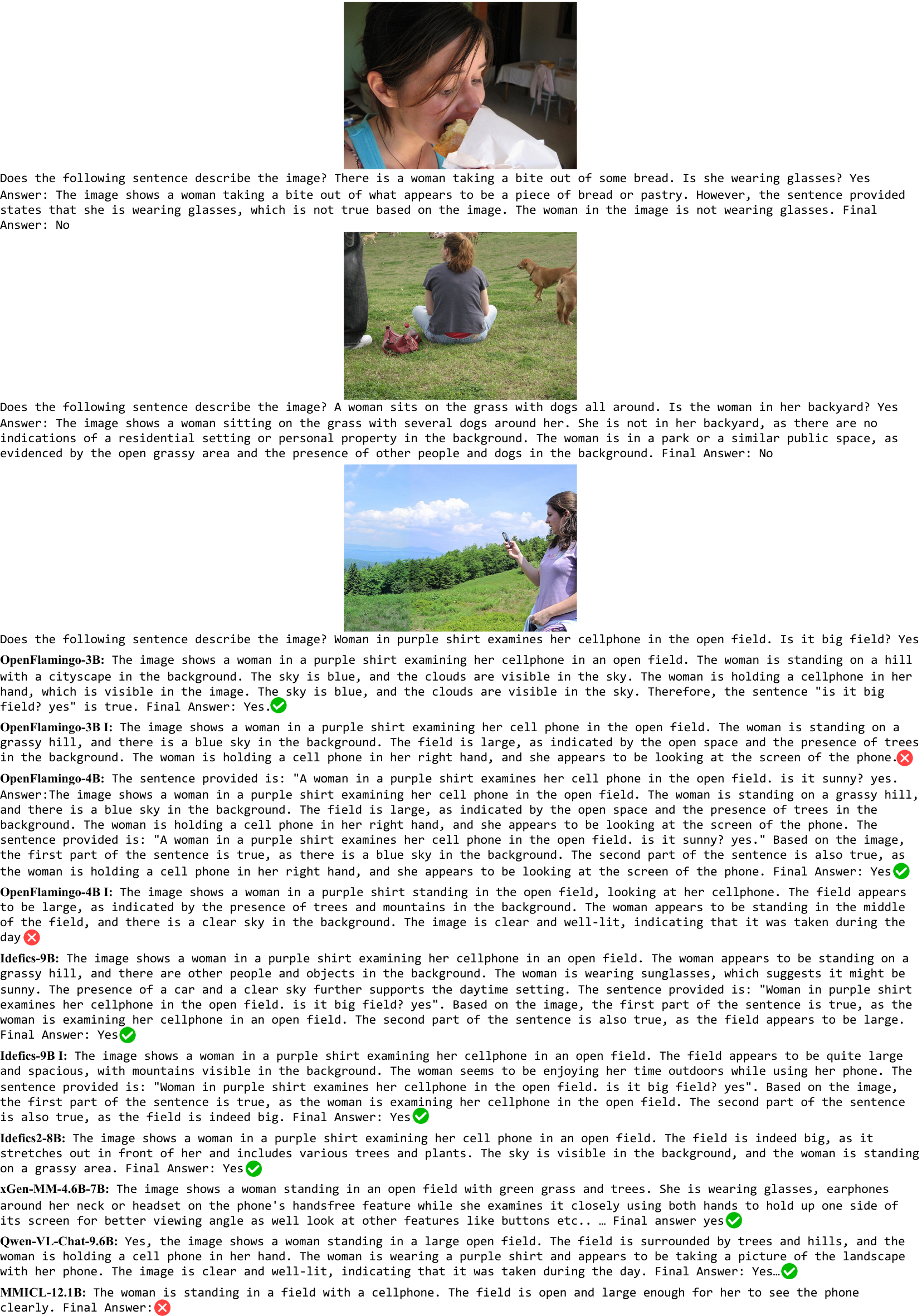}
    \caption{Example model predictions on instances from the \emph{Coreference} task, where demonstrations are selected based on visual and textual similarity, and Chain-of-Thought (CoT) reasoning is employed (setting \textbf{S+C}).}
    \label{fig:cot-coreference-detailed-examples}
\end{figure*}

\begin{figure*}[!t]
\centering
    \includegraphics[width=\linewidth]{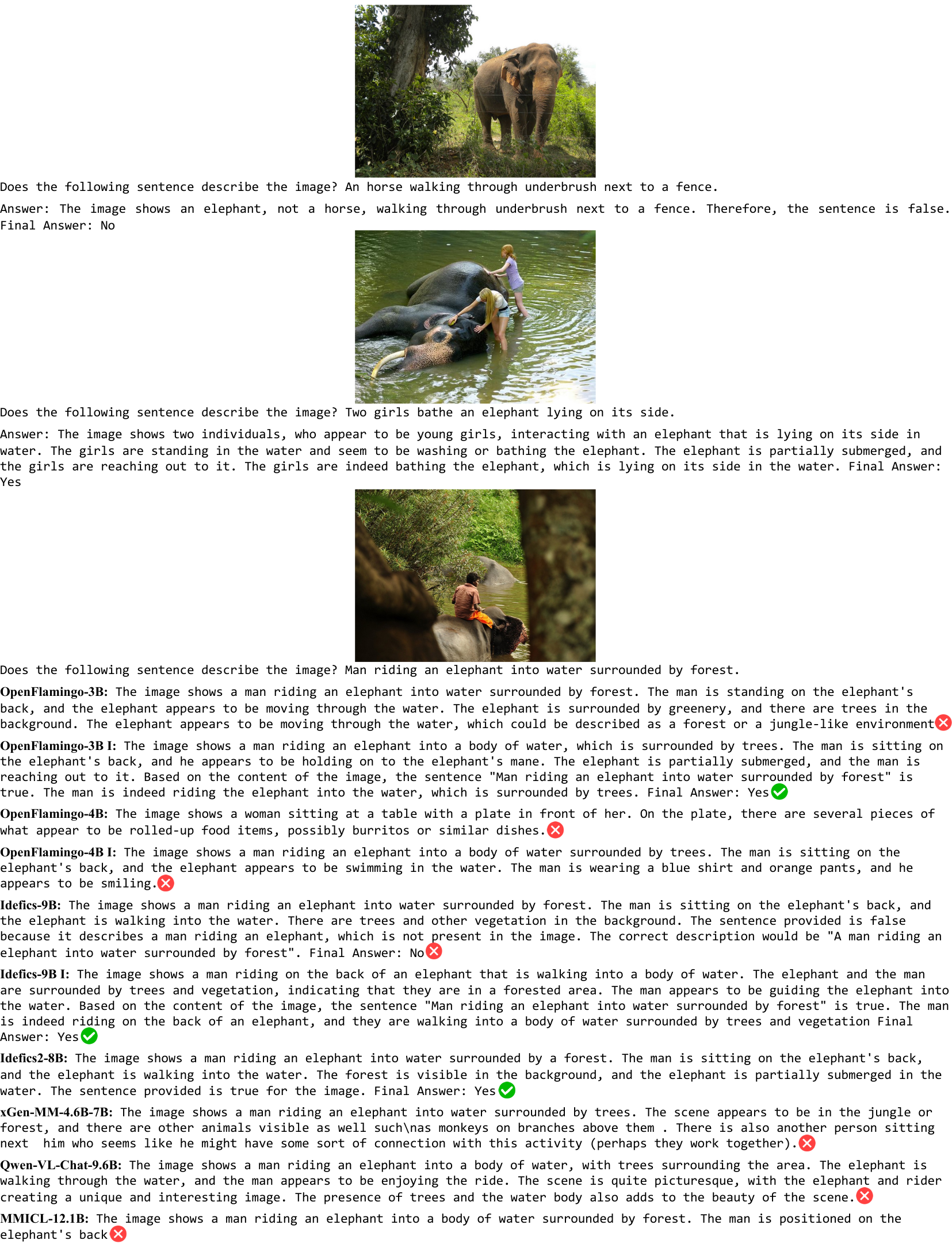}
    \caption{Example model predictions on instances from the \emph{Foil-It!} task, where demonstrations are selected based on visual and textual similarity, and Chain-of-Thought (CoT) reasoning is employed (setting \textbf{S+C}).}
    \label{fig:cot-foilit-detailed-examples}
\end{figure*}

In this section, we show example model responses from our evaluation. Figure \ref{fig:icl-existence-detailed-examples}-\ref{fig:icl-foilit-detailed-examples} illustrate the process of evaluating In-Context Learning (ICL). In this setup, demonstration examples are selected based on their similarity to the query and are provided with their ground truth answers before presenting the actual query image-text pair to the model. Similarly, Figure \ref{fig:cot-existence-detailed-examples}-\ref{fig:cot-foilit-detailed-examples} demonstrate the evaluation of Chain-of-Thought (CoT) reasoning combined with ICL. In this setting, CoT descriptions are included with the demonstration examples. These detailed reasoning chains guide the model in making inferences for the query image-text pair.

\end{document}